\documentclass[pdflatex,sn-aps]{sn-jnl}

\usepackage{graphicx}%
\usepackage{multirow}%
\usepackage{amsmath,amssymb,amsfonts}%
\usepackage{amsthm}%
\usepackage[title]{appendix}%
\usepackage{xcolor}%
\usepackage{textcomp}%
\usepackage{manyfoot}%
\usepackage{booktabs}%
\usepackage{algorithm}%
\usepackage{algorithmicx}%
\usepackage{algpseudocode}%
\usepackage{listings}%
\usepackage{tabularx}
\usepackage{threeparttable} 
\usepackage{booktabs} 
\usepackage{geometry} 
\usepackage{float} 
\usepackage{lipsum}         
\usepackage{subfig}  
\usepackage{array}    
\usepackage{ragged2e}

\theoremstyle{thmstyleone}%
\theoremstyle{thmstyletwo}%

\theoremstyle{thmstylethree}%

\begin{document}
\title[Robust Yoga Pose Classification]{Integrating Skeleton Based Representations for Robust Yoga Pose Classification Using Deep Learning Models}


\author[1]{\fnm{Mohammed} \sur{Mohiuddin}}\email{mohiuddin2531@gmail.com}

\author*[1]{\fnm{Syed Mohammod Minhaz} \sur{Hossain}}\email{minhazpuccse@gmail.com}

\author[1]{\fnm{Sumaiya} \sur{Khanam}}\email{sumaiya.khanam01@gmail.com}

\author[1]{\fnm{Prionkar} \sur{Barua}}\email{prionkar3301@gmail.com}

\author[1]{\fnm{Aparup} \sur{Barua}}\email{aparupdrubha3@gmail.com}

\author[1]{\fnm{MD Tamim} \sur{Hossain}}\email{thossain3333@gmail.com}

\affil[1]{\orgdiv{Department of Computer Science and Engineering}, \orgname{Premier University}, \orgaddress{\city{Chattogram}, \postcode{4000},  \country{Bangladesh}}}



\abstract{
 {Yoga is a popular form of exercise worldwide due to its spiritual and physical health benefits, but incorrect postures can lead to injuries. Automated yoga pose classification has therefore gained importance to reduce reliance on expert practitioners. While human pose keypoint extraction models have shown high potential in action recognition, systematic benchmarking for yoga pose recognition remains limited, as prior works often focus solely on raw images or a single pose extraction model. In this study, we introduce a curated dataset, “Yoga-16”, which addresses limitations of existing datasets, and systematically evaluate three deep learning architectures—VGG16, ResNet50, and Xception—using three input modalities: direct images, MediaPipe Pose skeleton images, and YOLOv8 Pose skeleton images. Our experiments demonstrate that skeleton-based representations outperform raw image inputs, with the highest accuracy of 96.09\% achieved by VGG16 with MediaPipe Pose skeleton input. Additionally, we provide interpretability analysis using Grad-CAM, offering insights into model decision-making for yoga pose classification with cross validation analysis.}}


\maketitle
\section{Introduction}\label{sec1}
Yoga has an origin tracing back thousands of years in India. It has evolved from
a spiritual and cultural practice into a globally embraced discipline. Renowned for its holistic benefits that encompass physical fitness, mental clarity, and spiritual harmony, yoga has become a cornerstone of wellness in the modern world. Its adoption has transcended geographical, cultural, and demographic boundaries, reflecting its universal appeal and adaptability. As this ancient practice gains momentum worldwide, the need for innovative solutions to make yoga more accessible, engaging, and precise has grown significantly.

 {In recent years, advancements in technology, particularly in artificial intelligence and machine learning, have created innovative opportunities to enhance numerous other domains} \cite{b1,b2,b3,b4,b5,b6,b7,b8,b9,b10,b11,b12,b13,b14,b15,b16} {, specifically in multimodal posture recognition systems} \cite{b43, b44, b45, b46, b47, b48, b49}. { Automated yoga pose recognition and correction systems have emerged as promising tools to democratize access to yoga training, enable personalized feedback, and maintain motivation among practitioners. These systems aim to replicate or augment the guidance traditionally provided by yoga instructors, making high-quality training accessible even in remote or resource-limited settings as highlighted in existing studies of yoga pose classification and similar tasks} \cite{b38,b39,b40,b41,b42}.  {Such technologies can potentially revolutionize the way yoga is practiced, offering opportunities for deeper engagement and more accurate adherence to correct postures.}

However, developing reliable and efficient automated systems for the classification of yoga poses is fraught with challenges. Existing datasets are often plagued by limitations such as noisy data, low-resolution images, and inadequate diversity, which collectively hinder the training of robust models. Moreover, while skeleton-based representations have shown significant promise in related fields such as human pose estimation, their potential in yoga pose classification remains largely underutilized. Nuanced variations in yoga postures, which often involve subtle differences in limb angles and body alignment, demand sophisticated methodologies to capture and classify these intricate details accurately.


 {Recent research indicates a limited comparative evaluation of skeletonized versus non-skeletonized approaches in yoga pose classification. Skeleton-based methods, focusing on body joint and alignment structures, may excel in identifying pose-specific features.  State-of-the-art human pose keypoint detection models, such as Mediapipe Pose and YOLOv8 Pose, have demonstrated potential in human pose detection. However, their effectiveness compared to traditional raw image-based methods remains underexplored. A comprehensive comparative study is crucial for advancing the field and optimizing systems for real-world applications.} Existing benchmark datasets like Yoga-82 exhibit issues such as class imbalance. State-of-the-art approaches like YogaConvo2D have achieved high accuracy but lack real-world challenges like inter-class similarity in their datasets. \cite{b17,b18,b19,b20}

This study conducts a comparative analysis, focusing on yoga pose selection and representation, to evaluate the performance of pose classification systems, comparing deep learning models such as VGG16 \cite{b21}, ResNet50 \cite{b22}, and Xception \cite{b23}. The poses chosen for this research reflect a deliberate effort to capture the diverse movement patterns, levels of complexity, and fitness benefits of yoga. Key factors influencing pose selection include:

\begin{enumerate}
    \item[(i)] Diverse Movement Patterns: The selected poses span various categories, including standing poses (Chair Pose, Tree Pose), balance-oriented poses (Side Plank, Lord of the Dance), and flexibility-enhancing poses (Seated Forward Bend, Low Lunge). This variety ensures comprehensive coverage of yoga movements, which is essential for creating versatile datasets.

   \item[(ii)] Broad Representation of Yoga Categories: The chosen poses include both asymmetric (Warrior I, Tree Pose) and symmetric poses (Chair Pose, Goddess Pose), as well as static (Wide Angle Forward Bend) and dynamic poses (Plank variants). This balance ensures a holistic representation of yoga practice.
   \item[(iii)] Popularity and Recognizability: Popular poses like the Downward Facing Dog and Warrior series are included to enhance the models' generalizability across different yoga styles.
    \item[(iv)] Fitness and Health Focus: The poses engage various muscle groups and aspects of physical fitness, addressing strength (Side Plank, Locust Pose), flexibility (Low Lunge), balance (Tree Pose), and endurance (Chair Pose).
   \item[(v)] Variation in Pose Complexity: By including simple (Staff Pose) and complex poses (Warrior 3), the dataset accommodates a wide range of user abilities and supports the development of models that generalize well across difficulty levels.
 \end{enumerate}

In addition to pose diversity, the overlapping features among yoga poses present unique challenges for pose classification. These include -
\begin{enumerate}
    \item[(i)] Similar Geometric Alignments: Poses like Dolphin Plank and Side Plank share core engagement and torso alignment, differing mainly in limb orientation.

    \item[(ii)] Partial Limb Overlaps: Standing poses such as Tree Pose and Lord of the Dance Pose exhibit similarities in lower-body posture.

    \item[(iii)] Ambiguities in Transitional and Dynamic Poses: Transitional poses like Downward Facing Dog share structural elements with other poses, while dynamic variations introduce further complexity.

\end{enumerate}
By conducting this comprehensive comparison study between skeletonized and non-skeletonized methodologies, this research aims to contribute to the broader understanding of AI’s role in wellness technologies and provide a foundation for future innovations in automated yoga pose classification systems.
In response to these challenges, this research introduces the following.
\begin{enumerate}
    \item Introduced Yoga-16, a curated dataset of 16 yoga poses, providing a standardized benchmark for evaluating yoga pose classification methods.
    \item Developed and compared three input modalities—direct image input, MediaPipe Pose skeleton images, and YOLOv8 Pose skeleton images—to systematically assess model performance across different data representations.
    \item Evaluated three deep learning architectures—VGG16, ResNet50, and Xception—on the Yoga-16 dataset, quantifying accuracy and generalization for both skeleton-based and non-skeleton-based inputs.
    \item Demonstrated the relative effectiveness of skeleton-based representations, with VGG16 achieving the highest benchmark accuracy of 96.09\% using MediaPipe Pose inputs, highlighting their potential for digital fitness applications.
    \item Applied Grad-CAM visualizations to interpret model decisions, providing insights into feature relevance and model behavior, and establishing interpretability as an integral part of the benchmarking framework. Additionally, to demonstrate the generalizability of the model we have implemented K-fold cross validation. 

\end{enumerate}


 {The remainder of this paper is organized as follows: Section} \ref{sec2}  {reviews related work on yoga pose classification and deep learning approaches; Section} \ref{sec3}  {details the methodology employed in this study, including dataset creation and model training; Section} \ref{sec4}  {presents the evaluation results of the proposed classification models across various input modalities; Section} \ref{sec5}  {discusses key findings and insights, along with potential application areas; and finally, Section} \ref{sec6}  {concludes the paper and outlines directions for future research.}

\section{Background and Related Work}\label{sec2}

Yoga pose classification has achieved significant achievement in recent years \& the recent studies shows that the transfer learning technique and Convolutional Neural Network (CNN) algorithms have a substantial impact on human pose estimation, including yoga pose estimation. In this literature review, we have categorized the existing work into three categories: (a) direct image input or non-skeletonization approach as shown in Table \ref{tab:non_skeletonization_summary}, (b) skeletonization approach as shown in Table \ref{tab:skeletonization_summary}, and (c) transfer learning approach as shown in Table \ref{tab:transfer_learning_summary}. In the non-skeletonization approach, we have explored that only CNN architectures have been used. On the other hand, in the skeletonization approach, various CNN models such as ResNet50, ResNet101, and Xception have been combined with object detection models such as YOLO, MediaPipe, and OpenPose. A few works have also utilized the same model for skeletonization and classification. Therefore, in the transfer learning approach, transfer learning models have been employed for classification without keypoint extraction, which we call a non-skeletonization approach.

\subsection{\textit {Benchmark Work Using Direct Image Input or Non-Skeletonization Approach}}

Verma et al. \cite{b19} proposed a dataset of 28.4k images distributed among 82 yoga poses belonging to six super classes/categories. They introduced the concept of fine-grained hierarchical pose classification to prevent false and complex annotations. Additionally, the authors presented the classification accuracy of CNN architectures on their proposed dataset. To utilize the hierarchical labels, they also depicted several hierarchical variants of DenseNet with the the highest 79.35\%  accuracy. The dataset contains 64 (min) and 1133 (max) images per class, with an average of 347 images per class.

Imran et al. \cite{b24} experimented on the Yoga-82 dataset using deep learning and modified some pre-trained models, including MobileNet, MobileNetV2, ResNet50, ResNet101, and Xception. Among these, the Xception model achieved the best result, with an overall accuracy of 82.52\% for 82 classes and 92.42\% for 6 super classes. They used a total of 18,488 images from the Yoga-82 dataset. Xception model was customized for optimization. After training, they recompiled the model, setting previously non-trainable parameters to trainable. Finally, they used an ensemble model in combination with the Keras model and applied one-hot encoding before feeding the features into Random Forest, which improved accuracy for both 82 classes and 6 classes.

Byeon et al. \cite{b25} proposed an ensemble deep model for posture recognition in various home environments. They introduced an ensemble system called EIR2TNet, consisting of InceptionNetV2 with five types of preprocessing, which showed good average performance compared to other recombination methods and pre-trained CNNs. Other used ensembles were VGGNet, ResNet, DenseNet, InceptionResNet, and Xception; all the aforementioned were trained on pre-trained convolutional neural networks with five different kinds of preprocessing. The experiment was done using an ETRI-made dataset from Korea containing 51,000 images divided into ten postures. The best obtained accuracy was 95.34\% for the EIR2TNet system.

Liaqat et al. \cite{b26} proposed a hybrid model that combined machine learning and deep neural networks to detect postures. The researchers focused on CNN and LSTM in an innovative architecture to decide postures. Following this, they developed hybrid models based on Deep Learning such as 1D-CNN, 2D-CNN, LSTM, and BiLSTM, and machine learning like KNN, Naive Bayes, Decision Tree, LDA, QDA, and SVM. They utilized the benchmark dataset in this respect, which was a well-known one, associated with the human body, and derived from Galvanic Skin Response. In this dataset, it had five subjects classified into three classes each: standing, sitting, and walking. The architecture they came up with was better than deep learning and machine learning models with accuracy of more than 98\% using hybrid CNN model.

\begin{table*}[ht]
\centering
\caption{Summary of Related Works Using Non-Skeletonization Approach}
\label{tab:non_skeletonization_summary}
\begin{tiny}
\begin{tabularx}{\textwidth}{@{} l p{0.15\textwidth} p{0.18\textwidth} p{0.12\textwidth} p{0.2\textwidth} p{0.15\textwidth} @{}}
    \toprule
    References & Source of Dataset & Amount of Images & Data Augmentation & Best Model & Accuracy \\
    \midrule
    \cite{b19} & Online & 28,478 & No & DenseNet-169 & 91.44\% \\
    \cite{b24} & Yoga-82 & 18,488 & No & Xception & 92.42\% \& 82.52\% \\
    \cite{b25} & Custom dataset & 51,000 & Yes & Ensemble deep model & 95.60\% \\
    \cite{b26} & Human Body using Galvanic Skin Response & N/A & N/A & Hybrid CNN (RAW data) & 98.14\% \\
    \bottomrule
\end{tabularx}
\end{tiny}
\end{table*}

\subsection{\textit {Benchmark Work Using Skeletonization Approach}}

Garg et al. \cite{b20} identified five main types of yoga postures—Downward Dog, Goddess, Plank, Tree, and Warrior—through low-latency models and measured the performances of multiple deep-learning architectures with and without skeletonization. Skeletonized images were found to generally help improve classification accuracy, with top performances achieved on non-skeletonized images by VGG16 (95.6\%) and on skeletonized images by YogaConvo2d (99.62\%). In this work, the skeletonization was done by utilizing the MediaPipe model. There was a comparative analysis on the dataset of 1,551 images of yoga poses. Their study was constrained to having only five easily distinguishable classes, this could potentially reduce the complexity of the classification task itself and narrow the generalizability of the findings to a wider set of more diverse or complex datasets of yoga poses.

Wadhwa et al. \cite{b27} conducted a performance comparison between OpenPose, MediaPipe, and YOLO models for pose recognition and network layer detection of postures. To detect poses using YOLOv8, the Yoga-82 dataset was utilized as the primary dataset. The YOLO algorithm was modified to detect body points and trained to recognize various poses. Although the entire dataset was used, the training focused on detecting five specific poses (Down-Dog, Goddess, Plank, Warrior 2, and Tree). OpenPose was also trained to detect five different poses (Airplane Pose, Mountain Pose, Sitting Pose, Triangle Pose, and V-shape Pose), while MediaPipe was similarly trained. Their paper provides a comparison chart showing that MediaPipe surpasses OpenPose in accuracy across all poses. One limitation of their work was the limited number of images used to detect yoga poses.

Agrawal et al. \cite{b28} proposed a machine-learning technique for identifying yoga poses using a skeletonization process. In their approach, they used tf-pose to extract body points and use these as features to train their models. The dataset for this experiment consisted of 5,500 images, with 80\% (4,367 images) used for training and 20\% (1,092 images) for testing. The best model they identified was the Random Forest Classifier, achieving an accuracy of 99.04\%. Other machine learning techniques they tested included Logistic Regression, SVM, Decision Tree, Naive Bayes, and KNN. The yoga poses were detected based on the angles extracted from skeleton joints using the tf-pose estimation algorithm. The limitations of their research include the restricted range of yoga poses and not utilizing deep learning models for potentially better performance.

Anilkumar et al. \cite{b29} presented a pose-estimation-based yoga monitoring system that relies on skeletonization, using MediaPipe as the model for geometric analysis of the captured frames through a camera. The system compares the detected angles from the skeleton to the text data of angles in the dataset, corresponding to different yoga poses. This system focuses on estimating the position of human body parts and joints in an image. Currently, their dataset contains only five major yoga poses for providing real-time feedback on the screen. A limitation of their system is that it operates only in two-dimensional space, without depth data considered for calculations. Additionally, the base threshold was 5 degrees, below which the MediaPipe pose detection system became unstable.

\begin{table*}[ht]
\centering
\caption{Summary of Related Work Using Skeletonization Approach}
\label{tab:skeletonization_summary}
\begin{tiny}
\begin{tabularx}{\textwidth}{@{} l p{0.18\textwidth} p{0.18\textwidth} p{0.12\textwidth} p{0.20\textwidth} p{0.15\textwidth} @{}}
    \toprule
    References & Source of Dataset & Amount of Images & Data Augmentation & Best Model & Accuracy \\
    \midrule
    \cite{b20} & Online & 1,551 & Yes & YogaConvo2d & 99.62\% \\
    \cite{b27} & Yoga-82, Kaggle & 18,488, 1,551 & No & YOLO \& MediaPipe \& OpenPose & 92.50\% \& 91.80\% \& 85.94\% \\
    \cite{b28} & Custom dataset & 5,500 & Yes & Random Forest & 99.04\% \\
    \bottomrule
\end{tabularx}
\end{tiny}
\end{table*}

\subsection {\textit{Benchmark Work Using Transfer Learning Approach}}

Long et al. \cite{b30} used six different transfer learning models to select the optimal model for the classification task. Their study shows that the TL-MobileNet-DA model gives the best performance with an overall accuracy of 98.43\%, and their task was to recognize yoga postures in real-time. They used 14 different yoga postures (bridge posture, cat-cow posture, child posture, cobra posture, corpse posture, downward-facing-dog posture, sitting posture, extended side-angle posture, warrior 2 posture, warrior 1 posture). A total of 1, 120 images were collected using RGB cameras from eight participants. Data augmentation was applied to boost the accuracy of the results. In their training process, they began by using the weights of each model-based and froze the top layers of the pre-trained model to fine-tune it for their dataset. Their findings indicated overall accuracies of 94.90\% for VGG16, 94.90\% for VGG19, 98.43\% for MobileNet, 92.16\% for MobileNetV2, 91.76\% for InceptionV3, and 98.04\% for DenseNet201. Moreover, they used the MediPipe algorithm to identify incorrect yoga postures.

Bilal et al. \cite{b31} proposed a transfer learning-based efficient spatiotemporal human action recognition framework for long and overlapping action classes. The transfer learning techniques were used for deep feature extraction. They claimed that their framework achieved state-of-the-art performance in spatiotemporal Human Action Recognition for overlapping human actions in long visual data streams. In their work, they used the UCF-101 dataset where they grouped similar action classes or those with overlapping actions. Among 13,320 video clips, all are in 101 action classes and the number of color channels is 3. As their training process, they proposed 7 deep learning pipelines where their proposed system showed an average accuracy of 96.03\%, making it better than other methods.

Jose et al. \cite{b32} presented a deep learning approach for yoga asana identification using transfer learning. In their work, they used transfer learning. The dataset used here consists of 10 classes and a total of 700 images. Features were extracted using transfer learning to generate a feature set, which was later split into train and test datasets and passed to a Deep Neural Network (DNN). For feature extraction, VGG16 as a transfer learning method was used, and a custom DNN was employed for prediction. Therefore, using the VGG16 architecture and pretrained ImageNet, they achieved an accuracy of 82\%.

\begin{table*}[ht]
\caption{Summary of Some Related Work Using Transfer Learning Approach}\label{tab:transfer_learning_summary}
\begin{tiny}
\begin{tabularx}{\textwidth}{@{} l p{0.2\textwidth} p{0.15\textwidth} p{0.15\textwidth} p{0.15\textwidth} p{0.15\textwidth} @{}}
    \toprule
    References & Source of Dataset & Amount of Images & Data Augmentation & Best Model & Accuracy \\
    \midrule
    \cite{b30} & Custom & 1,120 & Yes & TL-MobileNet-DA & 98.43\% \\
    \cite{b31} & UCF-101 dataset & 13,320 & Yes & Transfer learning-based framework & 96.03\% \\
    \cite{b32} & Online dataset & 700 & No & Transfer learning-based CNN & 82.00\% \\
    \bottomrule
\end{tabularx}
\end{tiny}
\end{table*}


\section{Methodology}\label{sec3}

\begin{figure}[htbp]
\centering
\includegraphics[width=8.4cm]{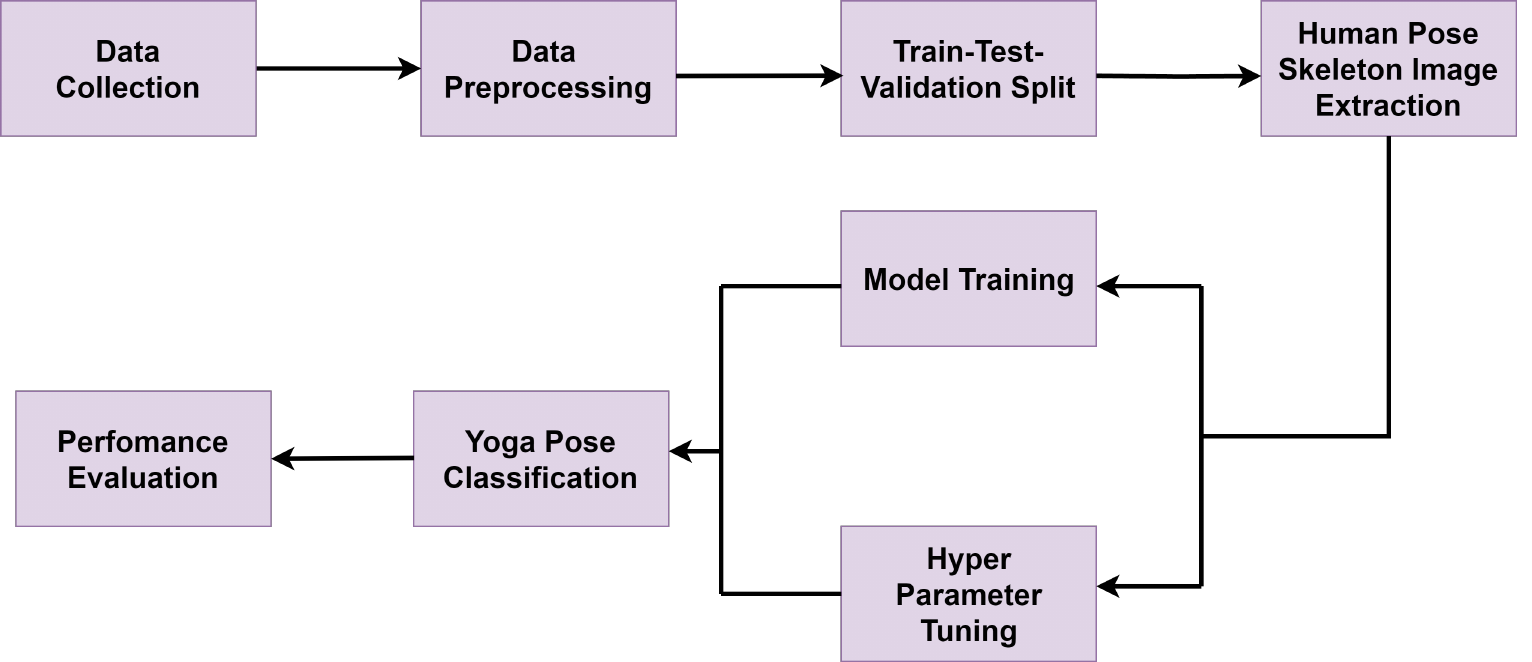}
\caption{The Workflow of Yoga Pose Classification} \label{workflow}
\end{figure}

This section presents the proposed methodology for yoga pose classification, as shown in Fig. \ref{workflow}. The data collection process, human pose keypoint extraction process, and training process of the deep learning architecture are also discussed.

\subsection{Problem Formulation}

The input to the model is an image labeled $X_i$ as shown in Eq. \eqref{eq:img}, where $h$, $w$, and $c$ represent the height, width, and number of channels (typically 3 for RGB images) of the input image, respectively.

\begin{equation}
    \text { $X_i \in \mathbb{R}^{h \times w \times c}$}
    \label{eq:img}
\end{equation}

We define a set of 16 classes as shown in Eq. \eqref{eq:classes}, corresponding to the different yoga poses in the dataset. For each image $X_i$, we have a label vector $Y_i$ as shown in Eq. \eqref{eq:yi}, representing the one-hot encoded ground truth.

\begin{equation}
    \text { $C = \{c_1, c_2, \ldots, c_{16}\}$}
    \label{eq:classes}
\end{equation}

\begin{equation}
    \text { $Y_i \in \{0, 1\}^{16}$}
    \label{eq:yi}
\end{equation}

We aim to learn a mapping function as shown in Eq. \eqref{eq:map_func}, which outputs a probability distribution over the 16 yoga pose classes. This uses a neural network model that applies global average pooling followed by a softmax activation function in the final layer.

\begin{equation}
    \text { $f : X_i \rightarrow Y_i$}
    \label{eq:map_func}
\end{equation}

To train the model, we minimize the categorical cross-entropy loss function as shown in Eq. \eqref{eq:cat_loss}, where $f(X_i)_j$ is the predicted probability for class $j$, and $Y_{ij}$ is the true label for the $i$-th image.
\begin{equation}
    L(f(X_i), Y_i) = - \sum_{j=1}^{16} Y_{ij} \log(f(X_i)_j)
    \label{eq:cat_loss}
\end{equation}

Across a dataset of $N$ yoga pose images, labelled $D$ as shown in Eq. \eqref{eq:dataset}, the objective is to minimize the average loss , which is mathematically shown in the Eq. \eqref{eq:avg_loss}. To achieve the goal of accurately predicting the correct yoga pose for each input image.

\begin{equation}
    \text { $D = \{(X_i, Y_i)\}_{i=1}^{N}$}
    \label{eq:dataset}
\end{equation}

\begin{equation}
    J(f) = \frac{1}{N} \sum_{i=1}^{N} L(f(X_i), Y_i)
    \label{eq:avg_loss}
\end{equation}

\subsection{Data collection and pre-processing}
\begin{figure}[H]
    \centering
    \begin{minipage}{0.23\textwidth}
        \centering
        \includegraphics[width=\linewidth]{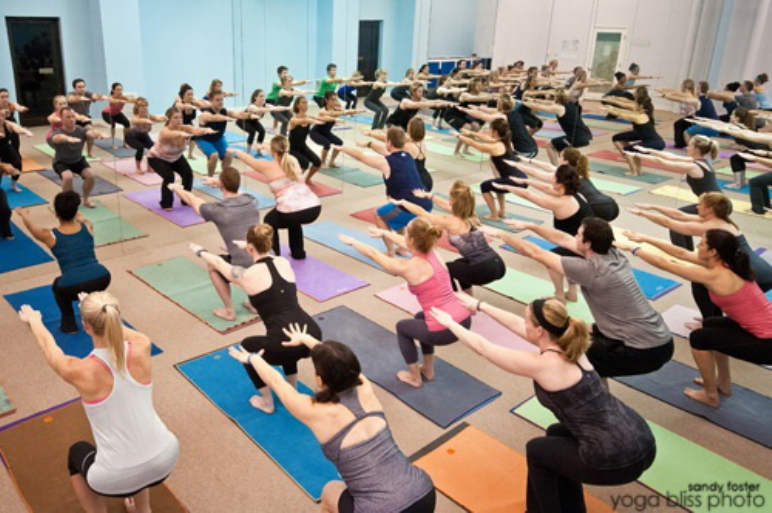}
        \\ (a) Multiple human subjects
    \end{minipage}
    \begin{minipage}{0.23\textwidth}
        \centering
        \includegraphics[width=\linewidth]{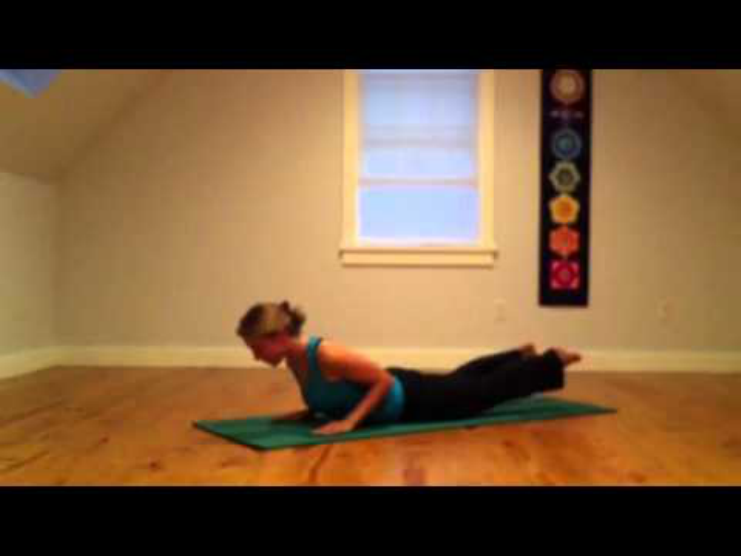}
        \\ (b) Low-quality image
    \end{minipage}
    \begin{minipage}{0.23\textwidth}
        \centering
        \includegraphics[width=\linewidth]{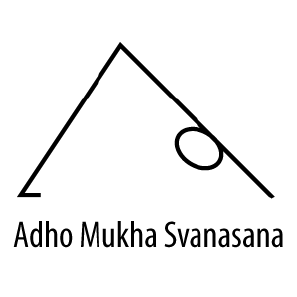}
        \\ (c) Non-human subjects
    \end{minipage}
    \begin{minipage}{0.23\textwidth}
        \centering
        \includegraphics[width=\linewidth]{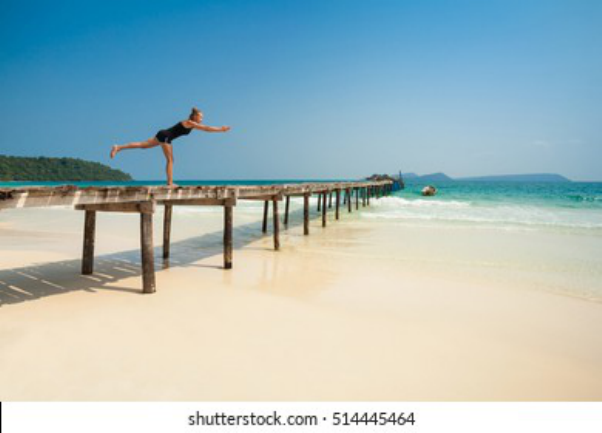}
        \\ (d) Zoomed-out subject
    \end{minipage}
    
    \caption{Examples of issues observed in the Yoga-82 dataset: (a) multiple human subjects in one image, (b) low-quality image, (c) non-human subjects such as shapes, and (d) zoomed-out individual makes identifying the pose difficult.}
    \label{fig:dataset_issues}
\end{figure}

We compiled images from the Yoga-82 dataset \cite{b19} and the Yoga Poses Dataset \cite{b33} to create a new dataset, ``Yoga-16'' \cite{b34}.  However, these datasets, along with other publicly available resources, revealed several issues that needed to be addressed before training the classification models. Notably, the Yoga-82 dataset \cite{b19} did not include the Goddess Pose class, which we supplemented using data from a publicly available Kaggle dataset \cite{b33}. Examples of these issues are illustrated in Fig. \ref{fig:dataset_issues}(a–d). 

The common challenges identified are as follows:

\begin{enumerate}
    \item \textbf{Subject Consistency:} The dataset was curated to ensure that each image contains a single human subject performing specific poses. This helps standardize the input for the model, providing consistent training examples from multiple subjects without ambiguity.

    \item \textbf{Relevance of Data:} We ensured that the dataset only contained images relevant to the task, specifically focusing on human poses. Non-relevant images, such as those containing non-human subjects (e.g., shapes or objects), were excluded to maintain focus on the primary objective.

    \item \textbf{High-Quality Image Selection:} To enhance the overall performance of the model, we removed low-quality images, such as those that were blurry, poorly lit, or unclear, to ensure that the model learns from well-defined and consistent data.

    \item \textbf{Image Cropping for Clarity:} We applied cropping techniques to highlight poses better, focusing on individual acting, thus improving the model's ability to recognize specific postures.

    \item \textbf{
Balanced Class Distribution:} We ensured a balanced representation across all classes by standardizing the number of images per class to approximately 80. This prevents any class from being overrepresented, which could lead to biased predictions, and helps the model generalize better.

\end{enumerate}

\begin{figure}[H]
    \centering
    \includegraphics[width=7cm]{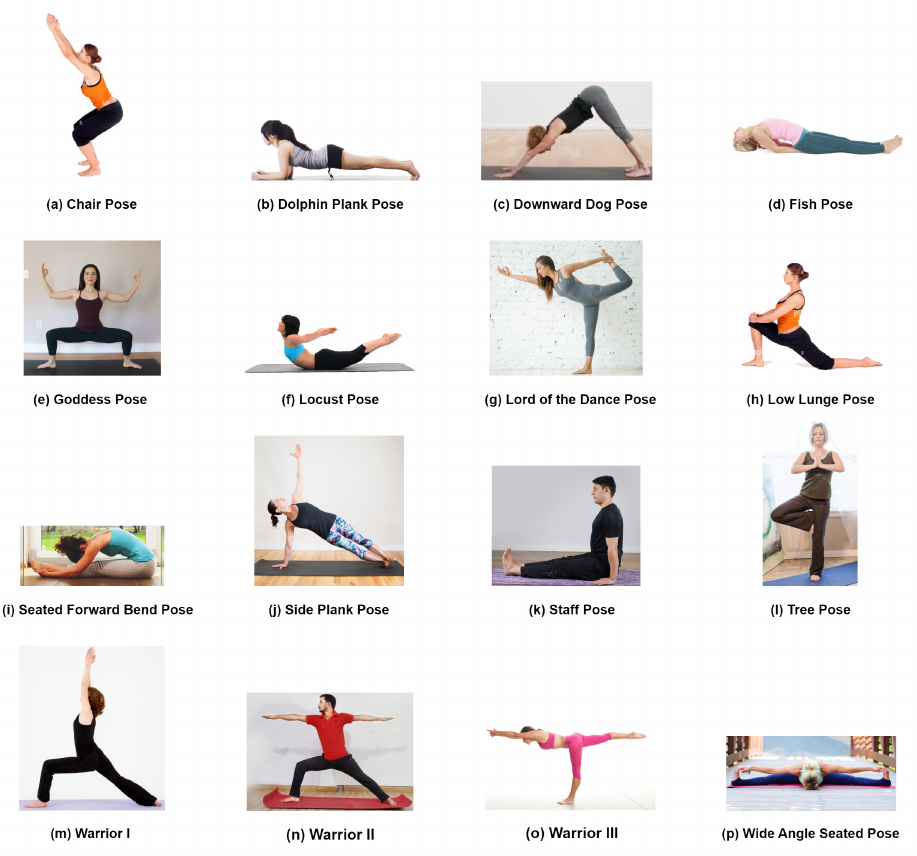}
    \caption{Sample images from Yoga-16 Dataset}
    \label{fig:sample_data}
\end{figure}


\begin{table}[h!]
\centering
\caption{
 {Summary of Lighting Variation and Camera Angle (Aspect Ratio) in Yoga-16 Dataset}
}
\label{tab:yoga_16_stats}
\begin{tabular}{lcccc}
 ine
\textbf{Metric} & \textbf{Mean} & \textbf{Std} & \textbf{Min} & \textbf{Max} \\
 ine
Brightness & 169.86 & 48.02 & 13.00 & 249.05 \\
Contrast & 59.34 & 14.94 & 20.45 & 117.59 \\
Aspect Ratio & 1.37 & 0.64 & 0.40 & 4.23 \\
 ine
\end{tabular}
\end{table}

In the process of collecting and preprocessing data, we focused on strategies that avoid introducing bias and ensure data quality and balance for model training. Table \ref{tab:yoga_16_stats}  {highlights the lighting and camera angle variations in the Yoga-16 dataset. Differences in brightness, contrast, and aspect ratio introduce diversity and realism, making the dataset more challenging and suitable for testing the generalization of pose classification models. As, yoga is more popular among women compared to men, most of the images contain women. }
Fig. ~\ref{fig:sample_data} illustrates a selection of preprocessed images from each class, and Table \ref{tab:dataset_overview} provides a detailed overview of the dataset.

\begin{table*}[h!]
\centering
\caption{Yoga Poses Dataset Overview}
\label{tab:dataset_overview}
\renewcommand{\arraystretch}{1.2}
\setlength{\tabcolsep}{3pt}
\begin{scriptsize}
\begin{tabularx}{\textwidth}{p{0.3\textwidth} p{0.18\textwidth} p{0.15\textwidth} *{3}{p{0.08\textwidth}}}
    \toprule
    Pose & Type & Source & Test & Train & Validation \\
    \midrule
    Chair & Standing & Yoga82 & 16 & 56 & 8 \\
    Dolphin Plank & Reclining & Yoga82 & 16 & 56 & 8 \\
    Downward Dog & Standing & Yoga82 & 16 & 56 & 8 \\
    Fish & Reclining & Yoga82 & 16 & 56 & 8 \\
    Goddess & Standing & Yoga Poses & 16 & 56 & 8 \\
    Locust & Reclining & Yoga82 & 16 & 56 & 8 \\
    Lord of Dance & Standing & Yoga82 & 16 & 56 & 8 \\
    Low Lunge & Standing & Yoga82 & 16 & 56 & 8 \\
    Seated Forward Bend & Sitting & Yoga82 & 16 & 56 & 8 \\
    Side Plank & Reclining & Yoga82 & 16 & 56 & 8 \\
    Staff & Sitting & Yoga82 & 16 & 56 & 8 \\
    Tree & Standing & Yoga82 & 16 & 56 & 8 \\
    Warrior 1 & Standing & Yoga82 & 16 & 56 & 8 \\
    Warrior 2 & Standing & Yoga82 & 16 & 56 & 8 \\
    Warrior 3 & Standing & Yoga82 & 16 & 56 & 8 \\
    Wide Angle Forward Bend & Sitting & Yoga82 & 16 & 56 & 8 \\
    \midrule
    \multicolumn{3}{l}{Total Images} & \textbf{256} & \textbf{896} & \textbf{128} \\
    \bottomrule
\end{tabularx}
\end{scriptsize}
\end{table*}


\subsection{Preliminaries}

 {
This subsection defines all key notations, operations, and formulas used across pose estimation and classification models. All repeated equations are consolidated here.
}
\subsubsection{Image Representation and Normalization}
Input images are represented as tensors:
\begin{equation}
    I \in \mathbb{R}^{H \times W \times C}
    \label{eq:image}
\end{equation}
where $H$ is height, $W$ is width, and $C$ is the number of channels. Pixel values are normalized to $[0,1]$:
\begin{equation}
    \text{Normalized pixel} = \frac{\text{Pixel Value}}{\text{Max Value}}
    \label{eq:normalized_pixel}
\end{equation}

\subsubsection{Core Deep Learning Operations}

\textbf{Convolution:}
\begin{equation}
    y[i,j,k] = \sum_{m,n,c} x[i+m-1,j+n-1,c] \cdot w[m,n,c,k] + b[k]
    \label{eq:conv}
\end{equation}

\textbf{Depthwise Separable Convolutions:}
\begin{equation}
    y[i,j,k] = \sum_{m=1}^{K} \sum_{n=1}^{K} x[i+m-1,j+n-1,k] \cdot w[m,n,k]
    \label{eq:depthwise}
\end{equation}
\begin{equation}
    z[i,j,p] = \sum_{k=1}^{C} y[i,j,k] \cdot v[k,p]
    \label{eq:pointwise}
\end{equation}

\textbf{Residual Connections:}
\begin{equation}
    y = f(x) + x
    \label{eq:residual}
\end{equation}

\textbf{ReLU Activation:}
\begin{equation}
    f(x) = \max(0, x)
    \label{eq:relu}
\end{equation}

\textbf{Max-Pooling:}
\begin{equation}
    y[i,j,k] = \max_{(m,n)\in[0,1]} x[2i+m, 2j+n, k]
    \label{eq:maxpool}
\end{equation}

\textbf{Global Average Pooling (GAP):}
\begin{equation}
    y_k = \frac{1}{H \times W} \sum_{i=1}^{H} \sum_{j=1}^{W} x[i,j,k]
    \label{eq:gap}
\end{equation}

\textbf{Softmax for Classification:}
\begin{equation}
    \sigma(z)_i = \frac{e^{z_i}}{\sum_{j=1}^{K} e^{z_j}}
    \label{eq:softmax}
\end{equation}

\subsubsection{Pose Keypoints and Skeleton Representation}

\textbf{Bounding Box:}
\begin{equation}
    B = (x, y, w, h)
    \label{eq:bb}
\end{equation}

\textbf{Heatmap-based Keypoint Detection:}
\begin{equation}
    P_{i,j} = \arg \max_{x,y} H_{i,j}(x,y)
    \label{eq:BP}
\end{equation}

\textbf{Non-Maximum Suppression (IoU):}
\begin{equation}
    \text{IoU}(A, B) = \frac{A \cap B}{A \cup B}
    \label{eq:iou}
\end{equation}

\textbf{Skeleton as Graph:}
\begin{equation}
    \mathcal{G} = (\mathcal{V}, \mathcal{E})
    \label{eq:graph}
\end{equation}

\subsubsection{Kalman Filter for Smoothing}

\textbf{Prediction:}
\begin{equation}
    \hat{x}_{k|k-1} = F \hat{x}_{k-1|k-1} + Bu_k
    \label{eq:pred1}
\end{equation}
\begin{equation}
    P_{k|k-1} = F P_{k-1|k-1} F^T + Q
    \label{eq:pred2}
\end{equation}

\textbf{Update:}
\begin{equation}
    K_k = P_{k|k-1} H^T (H P_{k|k-1} H^T + R)^{-1}
    \label{eq:up1}
\end{equation}
\begin{equation}
    \hat{x}_{k|k} = \hat{x}_{k|k-1} + K_k (z_k - H \hat{x}_{k|k-1})
    \label{eq:up2}
\end{equation}

\subsubsection{Training and Hyperparameter Tuning}

\textbf{Learning Rate Scheduling:}
\begin{equation}
    \eta_{t+1} = \eta_t \times \text{factor}
    \label{eq:lr_up}
\end{equation}

\textbf{Adam Optimizer Update:}

\begin{equation}
    \theta_{t+1} = \theta_t - \eta_t \cdot \frac{m_t}{\sqrt{v_t} + \epsilon}
    \label{eq:adam}
\end{equation}

\textbf{Categorical Cross-Entropy Loss:}
\begin{equation}
    L(y, \hat{y}) = - \sum_{i=1}^{N} y_i \log(\hat{y}_i)
    \label{eq:loss}
\end{equation}

\begin{figure}[htbp]
\centering
\includegraphics[width=8cm]{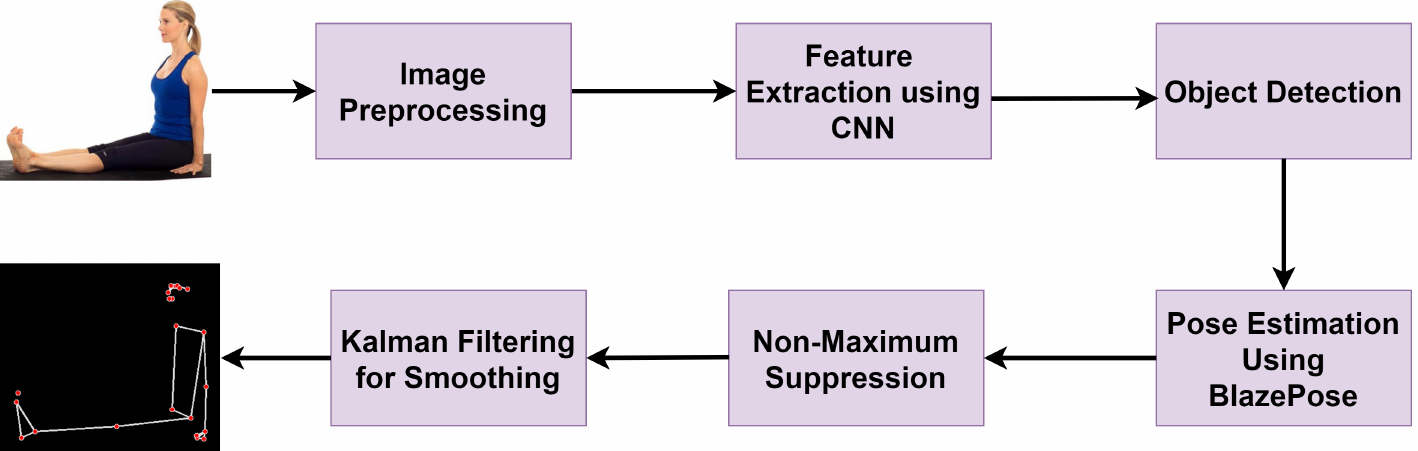}
\caption{Keypoint Extraction Pipeline in MediaPipe}
\label{fig:mediapipe_workflow}
\end{figure}

\subsection{Extracting Human Pose Keypoints using Mediapipe Model}
\begin{figure}[htbp]
\centering
\includegraphics[width=7cm]{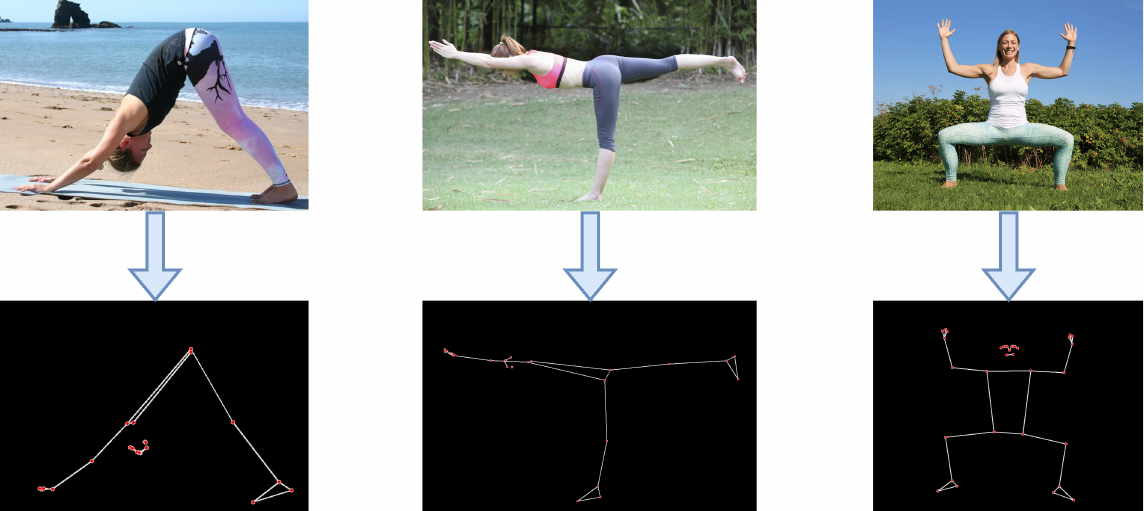}
\caption{Sample Output Images of MediaPipe Human Pose Skeleton Extraction}
\label{fig:mediapipe_skeleton}
\end{figure}

 {MediaPipe} \cite{b17}  {extracts human pose keypoints to generate simplified skeletons for classification. Images are preprocessed, features are extracted using CNNs, and objects are detected via bounding boxes. Pose estimation is performed with BlazePose} \cite{b35},  {followed by Non-Maximum Suppression to remove overlapping detections. Kalman filters smooth predictions (Eqs.}~\ref{eq:pred1} {--}\ref{eq:up2} {). Workflow and sample outputs are illustrated in Fig.} ~\ref{fig:mediapipe_workflow}  {and Fig.}~\ref{fig:mediapipe_skeleton} {. These processed skeletons are passed to downstream deep learning models for classification.}

\begin{figure}[htbp]
\centering
\includegraphics[width=8cm]{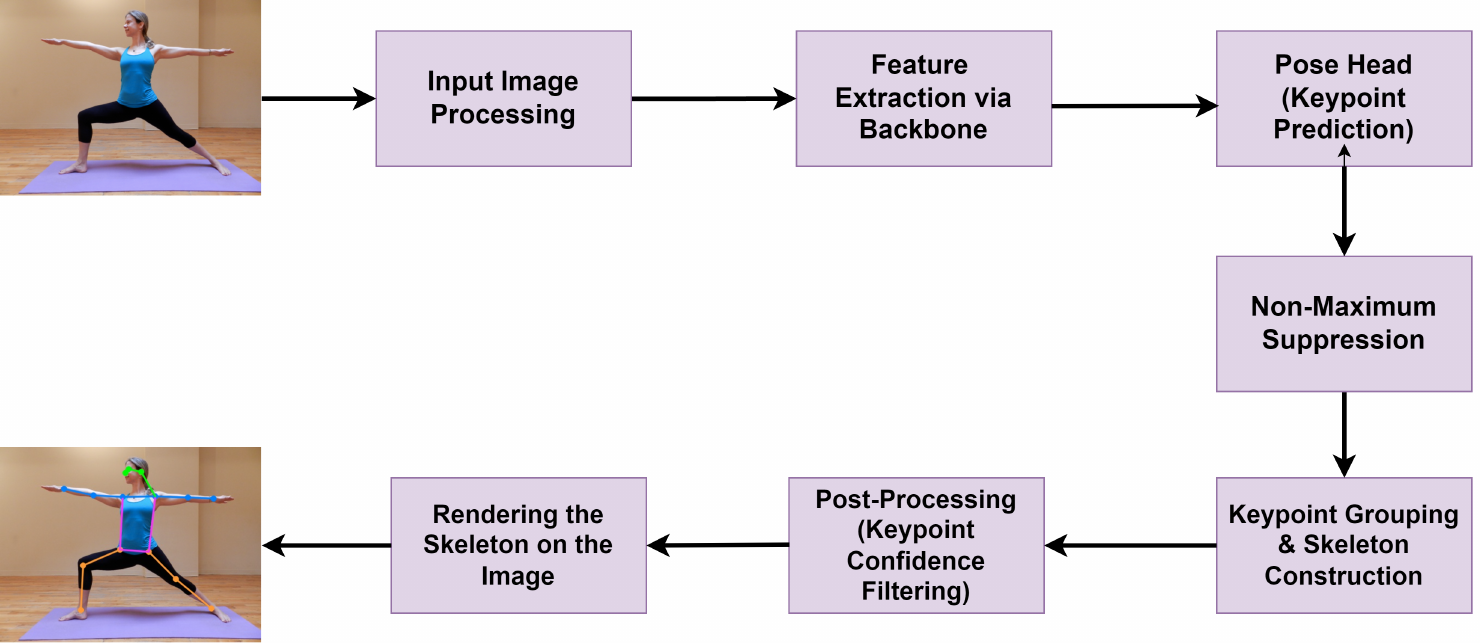}
\caption{Keypoint Extraction Pipeline in YOLOv8-Pose}
\label{fig:workflow_yolov8}
\end{figure}

\begin{figure}[htbp]
\centering
\includegraphics[width=6cm]{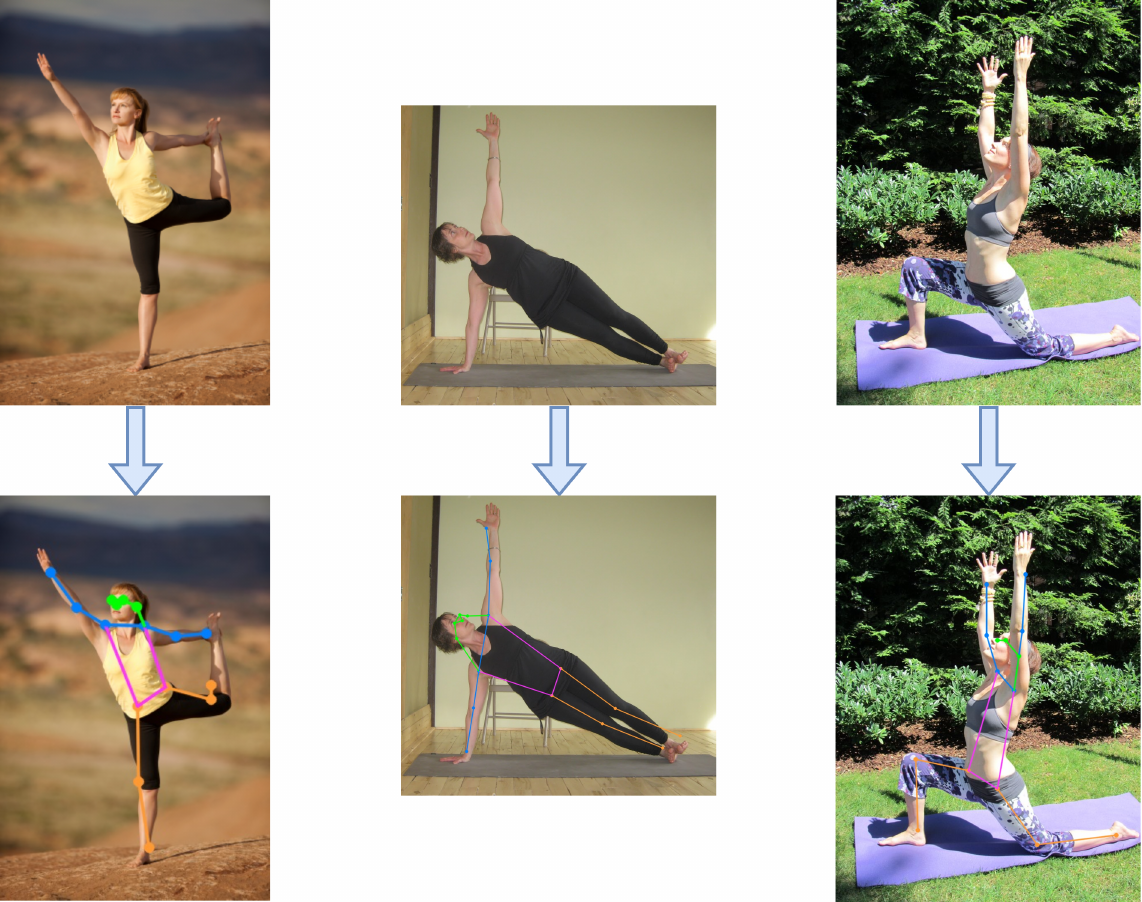}%
\caption{Sample Output Images of YOLOv8-Pose Keypoint Skeleton Extraction}
\label{fig:yolov8_samples}
\end{figure}

\subsection{Extracting Human Pose Keypoints using YOLOv8-Pose Model}
 {YOLOv8-Pose} \cite{b18}  {generates keypoint skeletons focusing on crucial joints while discarding irrelevant background. The input image} \(I\)  {is resized and normalized, then passed through a CNN backbone. The pose head predicts keypoints which are filtered via NMS (Eq.}~\ref{eq:iou} {) and grouped into a skeleton graph (Eq.}~\ref{eq:graph} {). Low-confidence keypoints are discarded, and the skeleton is rendered on the image. The workflow is shown in Fig.}~\ref{fig:workflow_yolov8}  {and the sample outputs are shown in Fig.}~\ref{fig:yolov8_samples}.

\begin{figure}[H] 
\centering
\includegraphics[width=\textwidth]{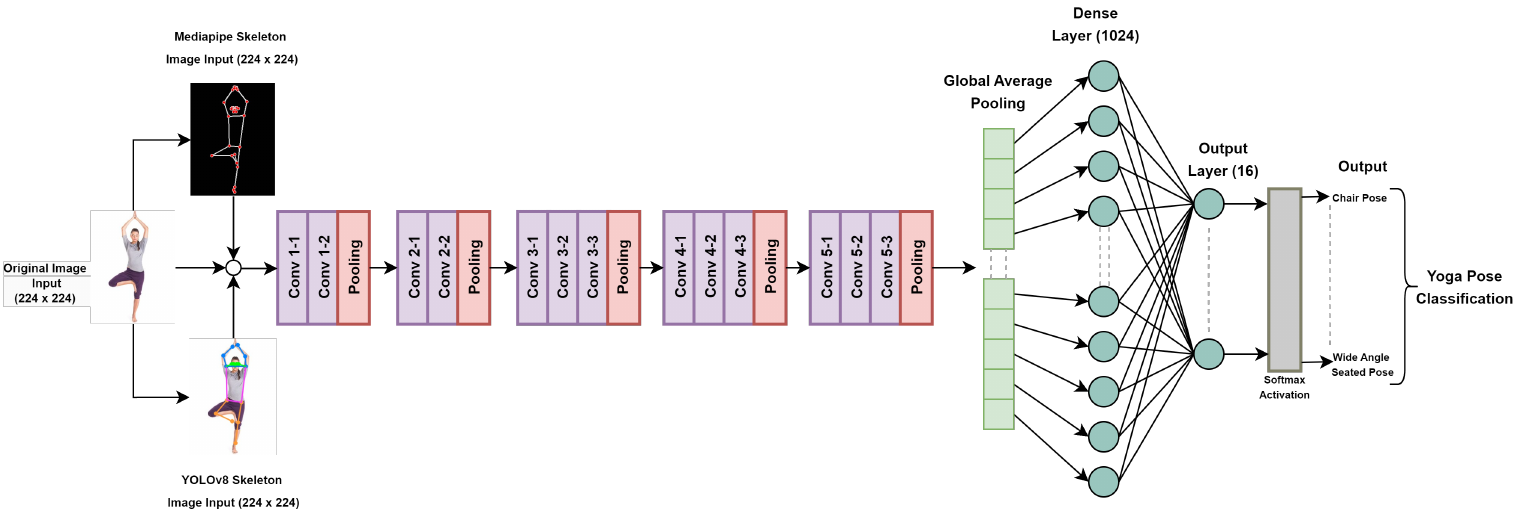}
\caption{VGG16 Architecture for Yoga Pose Classification Illustrating Flow of Input in Three Distinct Scenarios}
\label{fig:vgg16}
\end{figure}

\begin{figure}[H] 
\centering
\includegraphics[width=\textwidth]{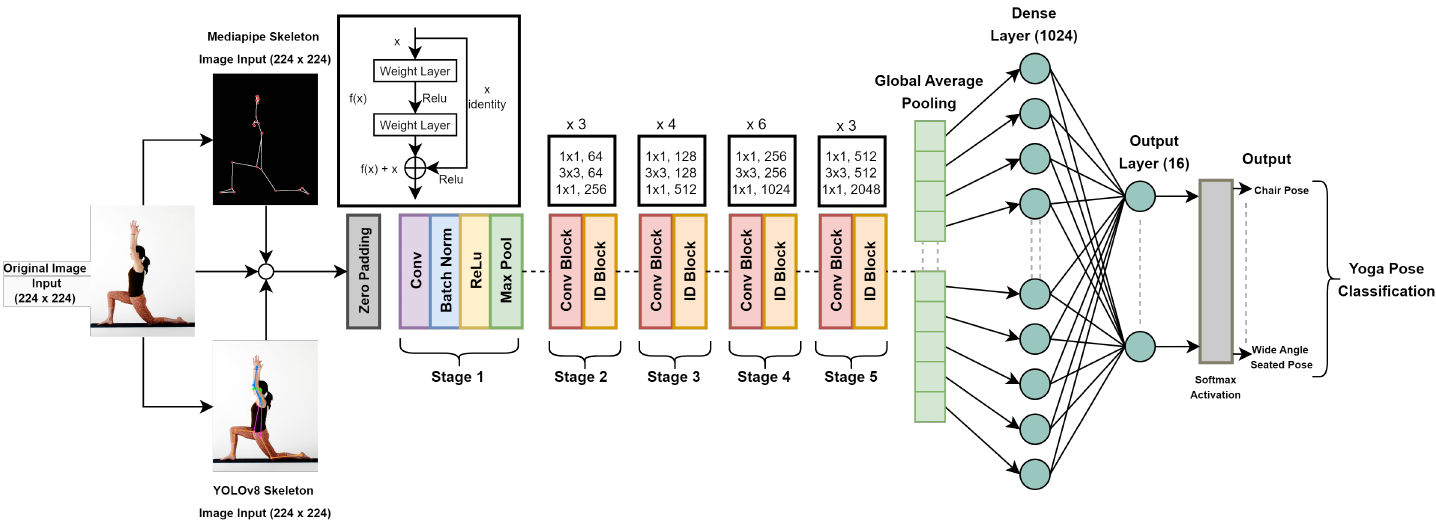}
\caption{ResNet50 Architecture for Yoga Pose Classification Illustrating Flow of Input in Three Distinct Scenarios}
\label{fig:ResNet50}
\end{figure}

\begin{figure}[H] 
\centering
\includegraphics[width=\textwidth]{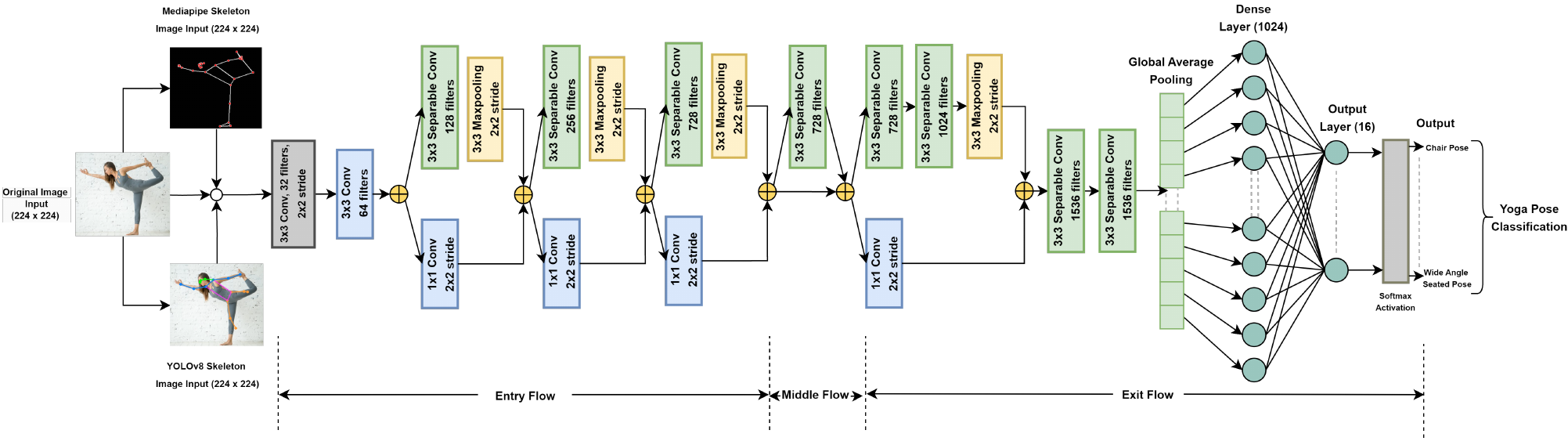}
\caption{Xception Architecture for Yoga Pose Classification Illustrating Flow of Input in Three Distinct Scenarios}
\label{fig:xception}
\end{figure}

\subsection{Classifying Yoga Poses}
 {VGG16, ResNet50, and Xception were employed for yoga pose classification, as illustrated in Fig.}~\ref{fig:vgg16}, Fig.~\ref{fig:ResNet50}, and Fig.~\ref{fig:xception}.  {All three models were applied to both non-skeleton images (direct inputs) and skeleton images. For both input types, features were extracted by training the CNN models from scratch, since pretrained ImageNet weights are not well aligned with skeletal heatmaps or pose-specific features as shown in Fig.}~\ref{fig:mediapipe_workflow} and ~\ref{fig:workflow_yolov8}.  {Training from scratch enabled the networks to learn task-specific representations directly from the Yoga-16 dataset, ensuring better adaptation to the unique characteristics of yoga poses. This enables the models to adapt its high-level representations of pose-specific skeletal patterns, improving discrimination between visually similar poses while retaining the benefits of transfer learning.}

\begin{table}[ht]
\centering
\caption{VGG16 Architecture for Yoga Pose Classification}
\label{tab:vgg16_architecture}
\footnotesize 
\begin{tabularx}{\textwidth}{@{\extracolsep{\fill}} l c c c c c @{}}
    \toprule
    \textbf{Function} & \textbf{Filter/Pool} & \textbf{\#Filters} & \textbf{Output} & \textbf{\#Parameters} & \textbf{Trainable} \\
    \midrule
    Input & - & - & $224 \times 224 \times 3$ & 0 & - \\
    Convolution & $3 \times 3$ & 64 & $224 \times 224 \times 64$ & 1,792 & Yes \\
    Convolution & $3 \times 3$ & 64 & $224 \times 224 \times 64$ & 36,928 & Yes \\
    Max Pooling & $2 \times 2$ & - & $112 \times 112 \times 64$ & 0 & - \\
    Convolution & $3 \times 3$ & 128 & $112 \times 112 \times 128$ & 73,856 & Yes \\
    Convolution & $3 \times 3$ & 128 & $112 \times 112 \times 128$ & 147,584 & Yes \\
    Max Pooling & $2 \times 2$ & - & $56 \times 56 \times 128$ & 0 & - \\
    Convolution & $3 \times 3$ & 256 & $56 \times 56 \times 256$ & 295,168 & Yes \\
    Convolution & $3 \times 3$ & 256 & $56 \times 56 \times 256$ & 590,080 & Yes \\
    Convolution & $3 \times 3$ & 256 & $56 \times 56 \times 256$ & 590,080 & Yes \\
    Max Pooling & $2 \times 2$ & - & $28 \times 28 \times 256$ & 0 & - \\
    Convolution & $3 \times 3$ & 512 & $28 \times 28 \times 512$ & 1,180,160 & Yes \\
    Convolution & $3 \times 3$ & 512 & $28 \times 28 \times 512$ & 2,359,808 & Yes \\
    Convolution & $3 \times 3$ & 512 & $28 \times 28 \times 512$ & 2,359,808 & Yes \\
    Max Pooling & $2 \times 2$ & - & $14 \times 14 \times 512$ & 0 & - \\
    Convolution & $3 \times 3$ & 512 & $14 \times 14 \times 512$ & 2,359,808 & Yes \\
    Convolution & $3 \times 3$ & 512 & $14 \times 14 \times 512$ & 2,359,808 & Yes \\
    Convolution & $3 \times 3$ & 512 & $14 \times 14 \times 512$ & 2,359,808 & Yes \\
    Max Pooling & $2 \times 2$ & - & $7 \times 7 \times 512$ & 0 & - \\
    Global Average Pooling & - & - & $1 \times 1 \times 512$ & 0 & Yes \\
    Dense & - & - & $1 \times 1024$ & 525,312 & Yes \\
    Dense & - & - & $1 \times 16$ & 16,400 & Yes \\
    \bottomrule
\end{tabularx}
\end{table}

\subsubsection{Classifying Yoga Poses using VGG16 Architecture}
 {VGG16} \cite{b21}  {is a deep neural network with a sequential design using small convolutional filters, enabling capture of intricate features as depth increases} (Table~\ref{tab:vgg16_architecture}).  {The input image} \(I\)  {with size 224×224×3 is processed through convolutional layers (Eq.}~\ref{eq:conv} {), ReLU activations (Eq.}~\ref{eq:relu} {), and max-pooling layers (Eq.}~\ref{eq:maxpool} {). The feature maps are then flattened, passed through dense layers, and classified with a softmax layer (Eq.}~\ref{eq:softmax} {). To reduce spatial dimensions while preserving discriminative features, Global Average Pooling is applied (Eq.}~\ref{eq:gap} {). These input images are then fed into VGG16, where fine-tuning is performed by unfreezing the last dense layers and partially the deeper convolutional blocks. }

\begin{table}[ht]
\centering
\caption{ResNet50 Architecture for Yoga Pose Classification}
\label{tab:resnet50_architecture}
\footnotesize 
\begin{tabularx}{\textwidth}{@{\extracolsep{\fill}} l m{2.5cm} c c c c @{}}
    \toprule
    \textbf{Layer} & \textbf{Type} & \textbf{Output Shape} & \textbf{\# Parameters} & \textbf{Trainable} \\
    \midrule
    Input & Input Layer & (224, 224, 3) & 0 & - \\
    Conv1 & 7x7 Convolution (64 filters) & (112, 112, 64) & $\sim 9.4K$ & No \\
    Max Pooling & 3x3 Max Pooling & (56, 56, 64) & 0 & No \\
    Stage 1 & Residual Block (3 blocks) & (56, 56, 256) & $\sim 75K$ & No \\
    Stage 2 & Residual Block (4 blocks) & (28, 28, 512) & $\sim 350K$ & No \\
    Stage 3 & Residual Block (6 blocks) & (14, 14, 1024) & $\sim 1.6M$ & No \\
    Stage 4 & Residual Block (3 blocks) & (7, 7, 2048) & $\sim 2.4M$ & No \\
    Global Average Pooling & GlobalAveragePooling2D & (1, 1, 2048) & 0 & Yes \\
    Dense (ReLU) & Dense Layer (1024 units) & (1024) & 2,098,176 & Yes \\
    Output Layer & Dense Layer (Softmax, 16 classes) & (16) & 16,400 & Yes \\
    \bottomrule
\end{tabularx}
\end{table}

\subsubsection{Classifying Yoga Poses using ResNet50 Architecture}
 {ResNet50} \cite{b22}  {leverages residual blocks to train deeper networks without vanishing gradients} (Table~\ref{tab:resnet50_architecture}).  {The input image} \(I\)  {with size 224×224×3 is processed through an initial convolution and max-pooling, followed by residual bottleneck blocks (Eq.}~\ref{eq:residual} {). ReLU activations (Eq.}~\ref{eq:relu} {) and Global Average Pooling (Eq.}~\ref{eq:gap} {) are applied before a dense layer with 1024 units and a softmax output (Eq.}~\ref{eq:softmax} {). These images are then processed by ResNet50 through unfreezing the last dense layers and partially the deeper convolutional blocks and tuned accordingly.}

\begin{table}[ht]
\centering
\caption{Xception Architecture for Yoga Pose Classification}
\label{tab:xception_architecture}
\footnotesize 
\begin{tabularx}{\textwidth}{@{\extracolsep{\fill}} l m{2.5cm} c c c c @{}}
    \toprule
    \textbf{Layer} & \textbf{Type} & \textbf{Output Shape} & \textbf{\# Parameters} & \textbf{Trainable} \\
    \midrule
    Input & Input Layer & (224, 224, 3) & 0 & No \\
    Conv1 & 3x3 Convolution (32 filters) & (112, 112, 32) & 896 & No \\
    Conv2 & 3x3 Convolution (64 filters) & (112, 112, 64) & 18,496 & No \\
    Max Pooling & 3x3 Max Pooling & (56, 56, 64) & 0 & No \\
    Entry Flow & Depthwise Separable Convolutions & (56, 56, 128) & $\sim 28K$ & No \\
    Middle Flow & Depthwise Separable Convolutions (8 blocks) & (28, 28, 728) & $\sim 1.7M$ & No \\
    Exit Flow & Depthwise Separable Convolutions & (14, 14, 2048) & $\sim 4.2M$ & No \\
    Global Average Pooling & GlobalAveragePooling2D & (1, 1, 2048) & 0 & Yes \\
    Dense (ReLU) & Dense Layer (1024 units) & (1024) & 2,098,176 & Yes \\
    Output Layer & Dense Layer (Softmax, 16 classes) & (16) & 16,400 & Yes \\
    \bottomrule
\end{tabularx}
\end{table}

\subsubsection{Classifying Yoga Poses using Xception Architecture}
 {Xception} \cite{b23}  {is built on depthwise separable convolutions for efficient feature learning} (Table~\ref{tab:xception_architecture}).  {The input image, } \(I\)  {with size 224×224×3 is processed through depthwise (Eq.}~\ref{eq:depthwise} {) and pointwise (Eq.}~\ref{eq:pointwise} {) convolutions, with ReLU activations (Eq.}~\ref{eq:relu} {) and residual connections (Eq.}~\ref{eq:residual} {). The resulting feature maps are reduced via Global Average Pooling (Eq.}~\ref{eq:gap} {) and passed through a dense layer with 1024 units, followed by a softmax output (Eq.}~\ref{eq:softmax} {) for classification.}

\begin{table}[ht]
\centering
\caption{Hyperparameters of Deep Learning Models used in the Yoga Pose Classification}
\label{tab:hyperparameters}
\begin{tabularx}{\textwidth}{@{\extracolsep{\fill}} l l @{}}
    \toprule
    \textbf{Hyperparameter} & \textbf{Values/Settings} \\
    \midrule
    Initial Learning Rate & 0.001 \\
    Learning Rate Reduction & 0.2 \\
    Minimum Learning Rate & 1e-6 \\
    Batch Size & 32 \\
    Optimizer & Adam with default parameters \\
    Epochs & 50 \\
    Early Stopping Patience & 10 \\
    Reduce Learning Rate Patience & 5 \\
    Loss Function & Categorical Cross-Entropy \\
    Model Checkpointing & Enabled \\
    \bottomrule
\end{tabularx}
\end{table}

\subsection{Hyperparameter Tuning for Deep Learning Models}

 {Training used learning rate scheduling} (Eq.~\ref{eq:lr_up})  {, Adam optimizer} (Eq.~\ref{eq:adam}) {, categorical cross-entropy loss} (Eq.~\ref{eq:loss}) {, batch size of 32, early stopping (patience 10), and model checkpointing} (Table~\ref{tab:hyperparameters}) {. These settings ensure stable and efficient training while preventing overfitting.}

\begin{table}[ht]
\centering
\caption{Experimental Settings Overview}
\label{tab:experiments_input_model}
\begin{tabularx}{\textwidth}{@{\extracolsep{\fill}} c c c @{}}
    \toprule
    \textbf{Experiment} & \textbf{Input Type} & \textbf{Model} \\
    \midrule
    1 & Direct Image & VGG16 \\
    2 & Direct Image & ResNet50 \\
    3 & Direct Image & Xception \\
    4 & Mediapipe Skeleton Image & VGG16 \\
    5 & Mediapipe Skeleton Image & ResNet50 \\
    6 & Mediapipe Skeleton Image & Xception \\
    7 & YOLOv8 Pose Skeleton Image & VGG16 \\
    8 & YOLOv8 Pose Skeleton Image & ResNet50 \\
    9 & YOLOv8 Pose Skeleton Image & Xception \\
    \bottomrule
\end{tabularx}
\end{table}

\begin{table}[ht]
\centering
\caption{System Configuration for Experiments}
\label{tab:sys_config}
\begin{tabularx}{\textwidth}{@{\extracolsep{\fill}} l X @{}}
    \toprule
    \textbf{Component} & \textbf{Specification} \\
    \midrule
    Platform & Kaggle \\
    GPU & NVIDIA Tesla P100 \\
    Processor & Intel(R) Xeon(R) CPU @ 2.00GHz \\
    RAM & 30 GB \\
    Disk Storage & 57.6 GB \\
    Operating System & Linux (Ubuntu) \\
    Frameworks and Libraries & TensorFlow, PyTorch \& Scikit-learn \\
    \bottomrule
\end{tabularx}
\end{table}


\section{Evaluation and Experimental Results}\label{sec4}
We have implemented 9 distinct yoga pose classification techniques as mentioned in Table \ref{tab:experiments_input_model}. All experiments were conducted on Kaggle’s cloud-based platform using consistent system configuration as shown in Table \ref{tab:sys_config}. The experiments were conducted in three input modalities - (i) Direct Image Input, (ii) Mediapipe Pose Skeleton Image Input and (iii) YOLOv8 Pose Skeleton Image Input, to evaluate and compare performance of skeleton based and non-skeleton based input scenarios. Under these input scenarios, we described the performance of deep learning models (VGG16, ResNet50 and  Xception)  below.


\begin{table}[h!]
\centering
\caption{ {Performance Assessment of Deep Learning Models with Different Input Types}}
\label{tab:combined_performance_assessment}
\begin{tabularx}{\textwidth}{@{\extracolsep{\fill}} l l c c c c @{}}
\toprule
\textbf{Input Type} & \textbf{Model} & \textbf{Accuracy (\%)} & \textbf{Precision (\%)} & \textbf{Recall (\%)} & \textbf{F1-Score (\%)} \\
\midrule
\multirow{3}{*}{Direct Image} 
    & VGG16     & \textbf{86.33} & \textbf{86.95} & \textbf{86.33} & \textbf{86.12} \\
    & ResNet50  & 66.41 & 66.76 & 66.41 & 66.08 \\
    & Xception  & 84.38 & 84.21 & 84.38 & 84.21 \\
\midrule
\multirow{3}{*}{Mediapipe Skeleton} 
    & VGG16     & \textbf{96.09} & \textbf{96.27} & \textbf{96.09} & \textbf{96.10} \\
    & ResNet50  & 88.28 & 89.04 & 88.28 & 88.16 \\
    & Xception  & 93.36 & 93.67 & 93.36 & 93.34 \\
\midrule
\multirow{3}{*}{YOLOv8-Pose Skeleton} 
    & VGG16     & \textbf{91.41} & \textbf{91.93} & \textbf{91.41} & \textbf{91.40} \\
    & ResNet50  & 75.00 & 76.42 & 75.00 & 74.97 \\
    & Xception  & 85.55 & 85.64 & 85.55 & 85.42 \\
\bottomrule
\end{tabularx}
\medskip

\textit{Note: Best results for each input type are shown in bold.}
\end{table}

\subsection{Performance Assessment of Deep Learning Models using Direct Image Input}

VGG16 achieves high accuracy 86.33\% as shown in Table \ref{tab:combined_performance_assessment} and excels at identifying specific poses like Chair Pose, Goddess Pose, and Seated Forward Bend Pose, with high precision and recall. However, it struggles with poses that have subtle structural features, such as Fish Pose. The confusion matrix, loss, and accuracy curves, and ROC curve of VGG16 with direct image input are shown in Fig. S1, Fig. S2, and Fig. S3, respectively. ResNet50, with an overall accuracy of 66.41\%, has lower precision and recall across many poses, particularly for Fish Pose and Warrior 2 Pose, indicating difficulty distinguishing similar poses. Despite these challenges, ResNet50 performs better with poses like Side Plank Pose and Warrior 3 Pose. The confusion matrix, loss, and accuracy curves, and ROC curve of ResNet50 with direct image input are shown in Fig. S4, Fig. S5, and Fig. S6, respectively. Xception, with an accuracy of 84.38\%, performs well across a wide range of poses, excelling in poses like Dolphin Plank Pose, Downward Facing Dog Pose, and Warrior 3 Pose. The confusion matrix, loss, and accuracy curves, and ROC curve of  Xception with direct image input are shown in Fig. S7, Fig. S8, and Fig. S9, respectively. Classification summaries for VGG16, ResNet50, and Xception are shown in  Table S1, Table S2, and Table S3.

\subsection{Performance of Pretrained Deep Learning Models on MediaPipe Skeleton Image Input}

MediaPipe skeleton extraction emphasizes key pose features, reducing background noise to enhance classification. VGG16 emerged as the best performer, achieving 96.09\% accuracy with balanced precision, recall, and F1-score, supported by consistent training progress and robust sensitivity-specificity balance. The confusion matrix, loss and accuracy curves, and ROC curve of VGG16 with Mediapipe skeleton image input are shown in Fig. S10, Fig. S11, and Fig. S12, respectively. ResNet50, while achieving 88.28\% accuracy, effectively captured skeletal details but faced occasional misclassifications, as reflected in its metrics and training curves. The confusion matrix, loss, and accuracy curves, and ROC curve of ResNet50 with Mediapipe skeleton image input are shown in Fig. S13, Fig. S14, and Fig. S15, respectively. Xception followed with 93.36\% accuracy, leveraging depthwise separable convolutions for robust classification despite minor errors. The confusion matrix, loss, and accuracy curves, and ROC curve of Xception with Mediapipe skeleton image input are shown in Fig. S16, Fig. S17, and Fig. S18, respectively. Classification summaries for VGG16, ResNet50, and Xception are shown in  Table \ref{tab:mediapipe_vgg16_classification_report}, Table S4, and Table S5. 

\subsection{Performance of Pretrained Deep Learning Models on YOLOv8-Pose Skeleton Image Input}

This section compares the performance of VGG16, ResNet50, and Xception on yoga pose classification using YOLOv8-Pose skeleton images, which emphasize key pose features by reducing background distractions. VGG16 achieved notable results with high accuracy, balanced sensitivity and specificity, and consistent training improvement. The confusion matrix, loss and accuracy curves, and ROC curve of Xception with YOLOv8 Pose skeleton image input are shown in Fig. S19, Fig. S20, and Fig. S21, respectively. ResNet50 demonstrated robust generalization with smooth training and a strong AUC, though it faced occasional misclassifications. The confusion matrix, loss and accuracy curves, and ROC curve of ResNet50 with YOLOv8 Pose skeleton image input are shown in Fig. S22, Fig. S23, and Fig. S24, respectively. Xception leveraged its efficient architecture to achieve high accuracy and low overfitting, despite minor challenges with similar classes. The confusion matrix, loss and accuracy curves, and ROC curve of Xception with YOLOv8 Pose skeleton image input are shown in Fig. S25, Fig. S26, and Fig. S27, respectively.  These results underline the models' effectiveness, with VGG16 and Xception excelling in this application. Classification summaries for VGG16, ResNet50, and Xception are shown in  Table S6, Table S7, and Table S8. 
In comparison, VGG16 achieved 91.41\%, ResNet50 achieved 75.00\%, and Xception achieved 85.55\% accuracy with YOLOv8 Pose skeleton image input as shown in Table \ref{tab:combined_performance_assessment}.

\begin{table}[ht]
\centering
\caption{Classification Report for VGG16 based classification framework using Mediapipe skeleton image input}
\label{tab:mediapipe_vgg16_classification_report}
\begin{tabularx}{\textwidth}{p{0.3\textwidth} p{0.16\textwidth} p{0.16\textwidth} p{0.13\textwidth} p{0.16\textwidth}}
    \toprule
    \textbf{Class Name} & \textbf{Accuracy (\%)} & \textbf{Precision (\%)} & \textbf{Recall (\%)} & \textbf{F1-Score (\%)} \\
    \midrule
    Chair Pose & 93.75 & 100.00 & 93.75 & 96.77 \\
    Dolphin Plank Pose & 93.75 & 100.00 & 93.75 & 96.77 \\
    Downward Facing Dog Pose & 100.00 & 100.00 & 100.00 & 100.00 \\
    Fish Pose & 87.50 & 87.50 & 87.50 & 87.50 \\
    Goddess Pose & 100.00 & 100.00 & 100.00 & 100.00 \\
    Locust Pose & 100.00 & 94.12 & 100.00 & 96.97 \\
    Lord of the Dance Pose & 93.75 & 100.00 & 93.75 & 96.77 \\
    Low Lunge Pose & 87.50 & 93.33 & 87.50 & 90.32 \\
    Seated Forward Bend Pose & 93.75 & 100.00 & 93.75 & 96.77 \\
    Side Plank Pose & 100.00 & 100.00 & 100.00 & 100.00 \\
    Staff Pose & 100.00 & 88.89 & 100.00 & 94.12 \\
    Tree Pose & 100.00 & 94.12 & 100.00 & 96.97 \\
    Warrior 1 Pose & 93.75 & 100.00 & 93.75 & 96.77 \\
    Warrior 2 Pose & 100.00 & 100.00 & 100.00 & 100.00 \\
    Warrior 3 Pose & 100.00 & 94.12 & 100.00 & 96.97 \\
    Wide Angle Seated Forward Bend Pose & 93.75 & 88.24 & 93.75 & 90.91 \\
    \midrule
    \textbf{Overall} & \textbf{96.09} & \textbf{96.27} & \textbf{96.09} & \textbf{96.10} \\
    \bottomrule
\end{tabularx}
\end{table}

\subsection{Best Model Performance: VGG16 with Mediapipe Skeleton Image Input}
From the comparison discussed in the previous section, it is observed that the highest performing model on this evaluation is VGG16 when using Mediapipe skeleton image input. As shown in Table \ref{tab:combined_performance_assessment}, we can observe that VGG16 performs better than the rest of the models, including ResNet50 and Xception, in all the evaluation metrics, with an accuracy of 96.09\%, a precision of 96.27\%, a recall of 96.09\% and an F1-score of 96.10\%.

This performance signifies that VGG16, when used with Mediapipe skeleton input, forms an excellent classier for the various poses in the dataset. The model's high performance is reflected in greater detail within the classification report found in Table \ref{tab:mediapipe_vgg16_classification_report}, in which the model achieves perfect recall and precision for many of the individual poses, such as the Downward Facing Dog Pose, Goddess Pose, and Warrior 2 Pose, to name but a few, with an overall accuracy of 96.09\%.
In all input types, the VGG16 model maintained the least number of trainable parameters: 541,712. At the same time, ResNet50 and Xception have a higher number of parameters: 2,114,576.  
The ability of VGG16 to consistently deliver better performance across a variety of metrics confirms it as the most reliable model in this work for skeleton-based pose classification using Mediapipe images.


\subsection{Hyperparameter Tuning for the best model: VGG16 with Mediapipe Skeleton Image Input}

A hyperparameter tuning process was carried out to improve the model’s performance further and reduce the risk of overfitting. This involved systematically adjusting key hyperparameters such as learning rate, batch size, number of epochs, and the optimizer type to find the optimal combination that would enhance generalization. Additionally, early stopping was employed to monitor the model’s performance on a validation set, halting training once the validation loss showed no improvement for a specified number of epochs. These measures helped ensure that the VGG16 model remained robust and generalizable, achieving excellent results without overfitting. VGG16's ability to consistently deliver better performance across a variety of metrics confirms it as the most reliable model for skeleton-based pose classification using Mediapipe images.

 {The performance of the VGG16 model for Mediapipe skeleton-based pose classification was enhanced through hyperparameter tuning, as summarized in Table} \ref{tab:vgg_hyperparameter}.  {Experiments showed that a} \(3 \times 3\)  {filter size consistently outperformed} \(5 \times 5\) {, likely because smaller filters are better at capturing fine-grained skeletal details, which are essential for distinguishing between similar poses. A pooling size of} \(2\times 2\)  {yielded higher accuracy and F1-scores compared to} \(3 \times 3\) {, as it preserves more spatial information, maintaining the relative positions of keypoints critical for accurate classification. Models trained with a batch size of 32 performed better than those with a batch size of 64, as smaller batches allow for more frequent and precise weight updates, improving the stability of gradient estimates and helping the model generalize better. Among optimizers, Adam consistently outperformed RMSProp due to its adaptive learning rate, which adjusts each parameter individually and allows the model to converge faster and more reliably. Furthermore, experiments with dense layer sizes revealed that a fully connected layer of 1024 neurons achieved better performance than smaller configurations such as 512 or 256. The larger dense layer provided higher representational capacity, enabling the network to model complex relationships between skeletal keypoints and capture subtle distinctions between challenging pose classes, while still avoiding overfitting due to proper regularization. The best configuration, achieved in Experiment 1, included a} \(3 \times 3\) filter size, \(2\times 2\)  {pooling size, batch size of 32, dense layer of 1024 neurons, and Adam optimizer, resulting in the highest accuracy (96.09\%) and F1-score (96.13\%). Overall, these hyperparameter choices helped the VGG16 model focus on subtle skeletal features, train efficiently, and generalize well, leading to superior performance in Mediapipe skeleton-based pose classification.}

\begin{table*}[ht] 
\centering
\caption{Hyperparameter Tuning Experiment of VGG16 with Mediapipe Skeleton Image Input (Best results are shown in bold)}
\resizebox{\textwidth}{!}{ 
\begin{tabular}{|c|c|c|c|c|c|c|c|c|c|c|}
 ine
\textbf{Exp. No.} & \textbf{Epochs} & \textbf{\#Filters} & \textbf{Filter Size} & \textbf{Pooling} & \textbf{Pooling Size} & \textbf{Optimizer} & \textbf{Batch Size} & \textbf{Dense Layer Size} & \textbf{Overall Accuracy (\%)} & \textbf{Overall F1-Score (\%)} \\
 ine
\textbf{1} & \textbf{100} & \textbf{64, 128, 256, 512} & \textbf{3x3} & \textbf{Max} & \textbf{2x2} & \textbf{Adam} & \textbf{32} & \textbf{1024} & \textbf{96.09} & \textbf{96.13} \\
2 & 100 & 64, 128, 256, 512 & 5x5 & Max & 2x2 & Adam & 32 & 512 & 95.01 & 94.50 \\
3 & 100 & 64, 128, 256, 512 & 3x3 & Max & 3x3 & Adam & 64 & 256 & 94.88 & 95.80 \\
4 & 100 & 64, 128, 256, 512 & 5x5 & Max & 3x3 & Adam & 64 & 1024 & 94.32 & 93.85 \\
5 & 100 & 64, 128, 256, 512 & 3x3 & Max & 2x2 & RMSProp & 32 & 512 & 93.19 & 94.12 \\
6 & 100 & 64, 128, 256, 512 & 5x5 & Max & 2x2 & RMSProp & 32 & 256 & 94.45 & 94.00 \\
7 & 100 & 64, 128, 256, 512 & 3x3 & Max & 3x3 & RMSProp & 64 & 1024 & 94.90 & 95.20 \\
8 & 100 & 64, 128, 256, 512 & 5x5 & Max & 3x3 & RMSProp & 64 & 512 & 95.12 & 92.65 \\
 ine
\end{tabular}
}
\label{tab:vgg_hyperparameter}
\end{table*}


\subsection{Evaluation of Generalization for Our Best Model: VGG16 with Mediapipe Skeleton Image Input}

 {To validate the robustness of the Mediapipe+VGG16 model on a limited dataset, we applied 5-fold cross-validation. The results ($93.55 \pm 0.94$) summarized in }Table \ref{tab:perfold-metrics}  {were slightly lower than the base model accuracy of 96.09\%, as expected due to evaluation across multiple, diverse data splits. The low standard deviation demonstrates consistent performance across folds, confirming the model’s reliability and strong generalization capability.

To comprehensively evaluate the generalization capability of the Mediapipe+VGG16 model, we conducted experiments using a custom dataset} \cite{b36} {specifically designed to simulate real-world scenarios. This dataset was constructed by capturing 20 images per yoga pose class collected from YouTube videos, which introduced significant diversity in pose presentation and background elements. These variations were intended to test the model's ability to perform accurately under conditions that differ substantially from the controlled environment of the original dataset.}


\begin{table}[h]
\centering
\caption{ {Per-fold performance metrics with mean and standard deviation}}
\label{tab:perfold-metrics}
\begin{tabular}{c|c|c|c|c}
 ine
\textbf{Fold} & \textbf{Accuracy} & \textbf{Precision} & \textbf{Recall} & \textbf{F1-score} \\
 ine
1 & 0.921951 & 0.927367 & 0.921951 & 0.922230 \\
2 & 0.946341 & 0.949426 & 0.946341 & 0.946471 \\
3 & 0.941463 & 0.944789 & 0.941463 & 0.942095 \\
4 & 0.936585 & 0.944359 & 0.936585 & 0.935641 \\
5 & 0.931373 & 0.931566 & 0.931373 & 0.929679 \\
 ine
\textbf{Mean $\pm$ SD} & 
$93.55 \pm 0.94$\% & 
$93.95 \pm 0.95$\% & 
$93.55 \pm 0.94$\% & 
$93.52 \pm 0.97$\% \\
 ine
\end{tabular}
\end{table}

\begin{table}[htbp!]
    \centering
    \setlength{\tabcolsep}{5pt} 
    \renewcommand{\arraystretch}{1.2} 
    \caption{Evaluation of generalization}
    \label{tab:dataset_comparison}
    \begin{tabularx}{\columnwidth}{X cccc}
        \toprule
        \textbf{Dataset} & \textbf{Accuracy (\%)} & \textbf{Precision (\%)} & \textbf{Recall (\%)} & \textbf{F1-score (\%)} \\
        \midrule
        Our Dataset (\cite{b34}) & 96.09 & 96.27 & 96.09 & 96.10 \\
        Custom Test Set (\cite{b37}) & 93.75 & 94.41 & 93.75 & 93.78 \\
        \bottomrule
    \end{tabularx}
\end{table}

 {The results demonstrated the model's effectiveness in generalizing to previously unseen data, achieving an accuracy of 93.75\%, a precision of 94.41\%, a recall of 93.75\%, and an F1-score of 93.78\%. These metrics highlight the model's robustness, as it maintained high performance despite the added complexity and variability in the test set.
However, a slight performance drop was observed when compared to the scores obtained on the original dataset, as detailed in Table} \ref{tab:dataset_comparison}.  {Despite this, the model's scores remained strong, indicating that it can adapt effectively to data drawn from real-world sources. Overall,  This capability to generalize well beyond the training conditions enhances the model's reliability and broadens its applicability in real-world settings.}

\begin{table*}[htbp!]
    \centering
    \setlength{\tabcolsep}{2pt}  
    \renewcommand{\arraystretch}{1.3}  
    \caption{Summary of comparison between state-of-the-art approaches for yoga pose classification}
    \label{tab:soa_result_comparison}
    \begin{tiny}
    \begin{tabularx}{\textwidth}{p{0.1\textwidth} p{0.08\textwidth} p{0.12\textwidth} p{0.12\textwidth} p{0.12\textwidth} p{0.12\textwidth} p{0.12\textwidth} p{0.12\textwidth}}
        \toprule
        \textbf{Reference} & \textbf{\shortstack{No of\\Classes}} & \textbf{\shortstack{Low\\Resolution \\Dataset}} & \textbf{\shortstack{Class\\Imbalance}} & \textbf{\shortstack{Classes with\\Overlapped \\Features}} & \textbf{\shortstack{Skeletonized\\Image}} & \textbf{\shortstack{Non-\\Skeletonized\\Image}} & \textbf{\shortstack{Comparison\\ between\\Skeletonization\\ \& Non-\\Skeletonization}} \\  
        \midrule

        \cite{b19} & 82 & NR & NR & R & NR & R & NR \\
        \cite{b20} & 5 & NR & NR & NR & R & R & R \\
        \cite{b27} & 5 & NR & NR & NR & R & NR & R \\
        \cite{b24} & 6 & NR & NR & R & NR & R & NR \\
        \cite{b25} & 11 & R & R & R & NR & R & NR \\
        \cite{b28} & 10 & R & NR & NR & R & NR & NR \\
        \cite{b30} & 14 & NR & R & R & NR & R & NR \\
        \cite{b32} & 10 & R & R & R & NR & R & NR \\

        \textbf{Our Work} & 16 & R & R & R & R & R & R \\ 
        \bottomrule 
    \end{tabularx}
    \begin{flushleft}
        \textbf{NR = Not Resolved, R = Resolved}
    \end{flushleft}
    \end{tiny}
\end{table*}

\subsection{Comparison between State-of-the-Art Approaches for Yoga Pose Classification}

The comparison of the results by the existing state-of-the-art approaches is summarized in Table \ref{tab:soa_result_comparison}.
The proposed approach comprehensively addresses key challenges such as low-resolution datasets, class imbalance, overlapping class features, and evaluation using both skeletonized and non-skeletonized images. It offers a robust evaluation framework and demonstrates scalability by effectively classifying 16 different yoga poses.

By contrast, previous works are only partially solved with respect to the mentioned challenges: for example, \cite{b19} and \cite{b24} solve overlapping class features and use non-skeletonized images but fail to address other significant challenges such as low-resolution datasets or class imbalance. Moreover, \cite{b19} supports more classes - 82-while \cite{b24} only handles six classes. Similarly, \cite{b25} and \cite{b32} provide balanced solutions by addressing low-resolution datasets, class imbalance, and overlapping features, yet they do not incorporate skeletonized images into their methodologies.

Works like \cite{b20} and \cite{b27} lean more towards skeletonized image processing but fail to overcome challenges such as low-resolution datasets and overlapping class features. Also, \cite{b28} and \cite{b30} have partial solutions for some of the above challenges- low-resolution datasets or overlap in features—but lack a holistic evaluation approach.

The proposed approach outperforms existing methods by addressing key challenges comprehensively and integrating skeletonized and non-skeletonized representations. This dual strategy ensures robust, accurate yoga pose classification, offering a scalable and reliable solution for real-world applications.

\begin{figure}[htbp]
\centering
\includegraphics[width=8.5cm]{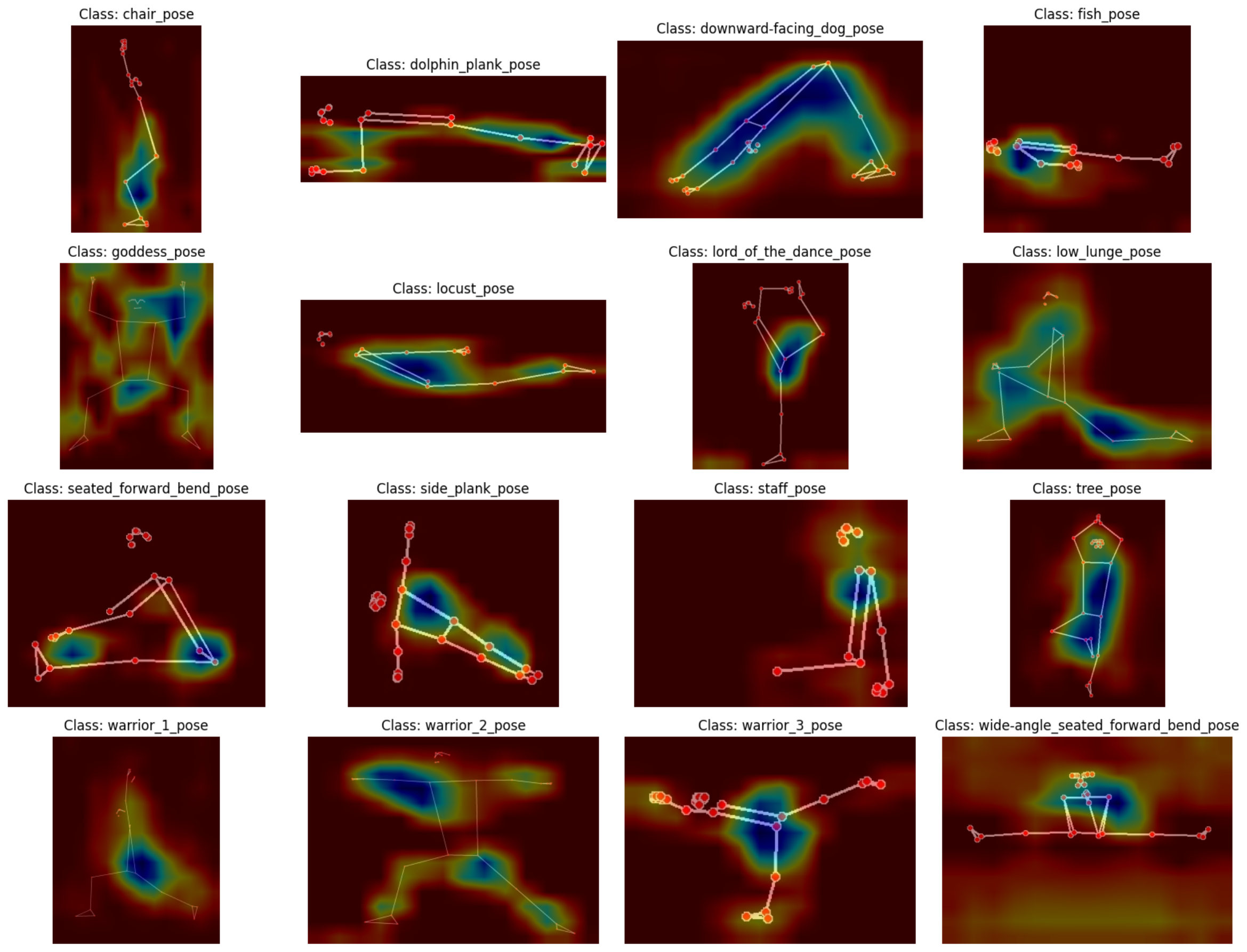}
\caption{Grad-CAM Visualization 16 Classified Poses using VGG16 Model using Mediapipe Skeleton Image Input.}
\label{fig:gradcam_vgg16}
\end{figure}

\begin{figure}[htbp]
    \centering
    \begin{tabular}{cc}
        \subfloat[\centering]
        {\includegraphics[width=4cm]{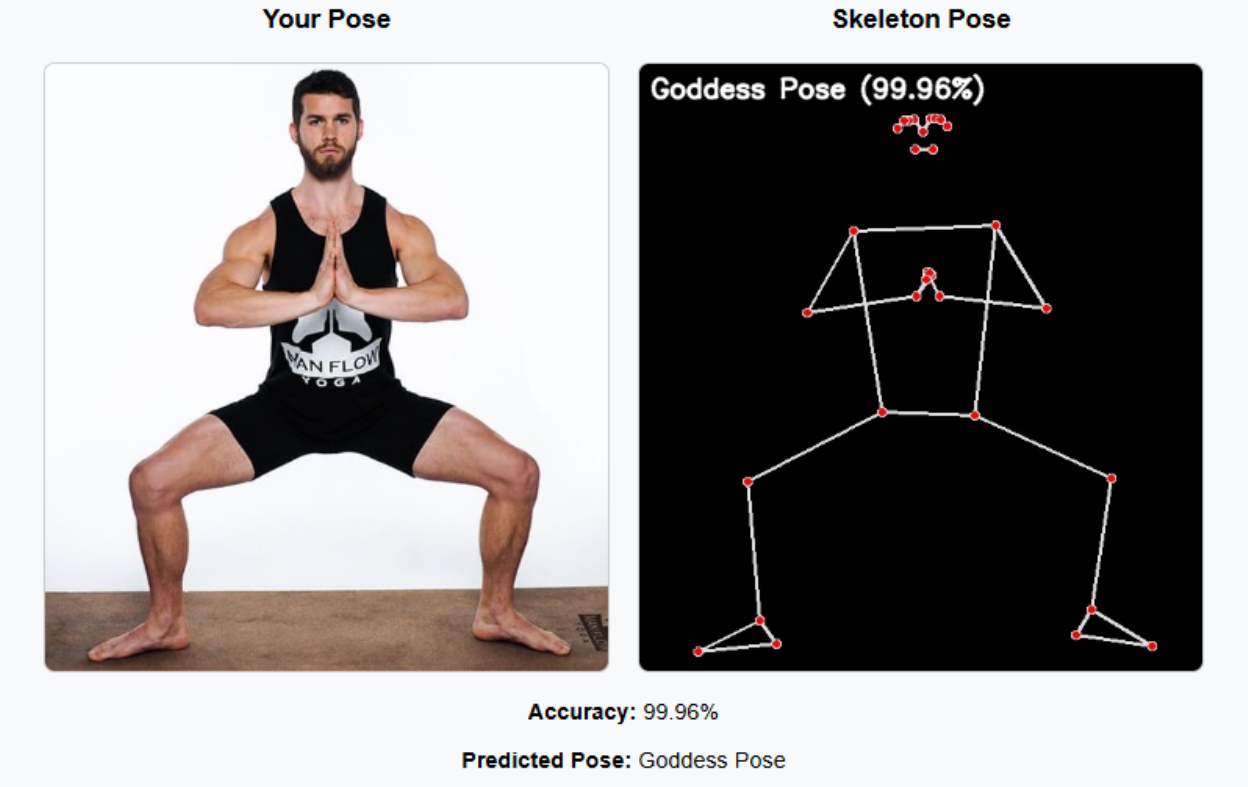}} &

        \subfloat[\centering]
        {\includegraphics[width=4cm]{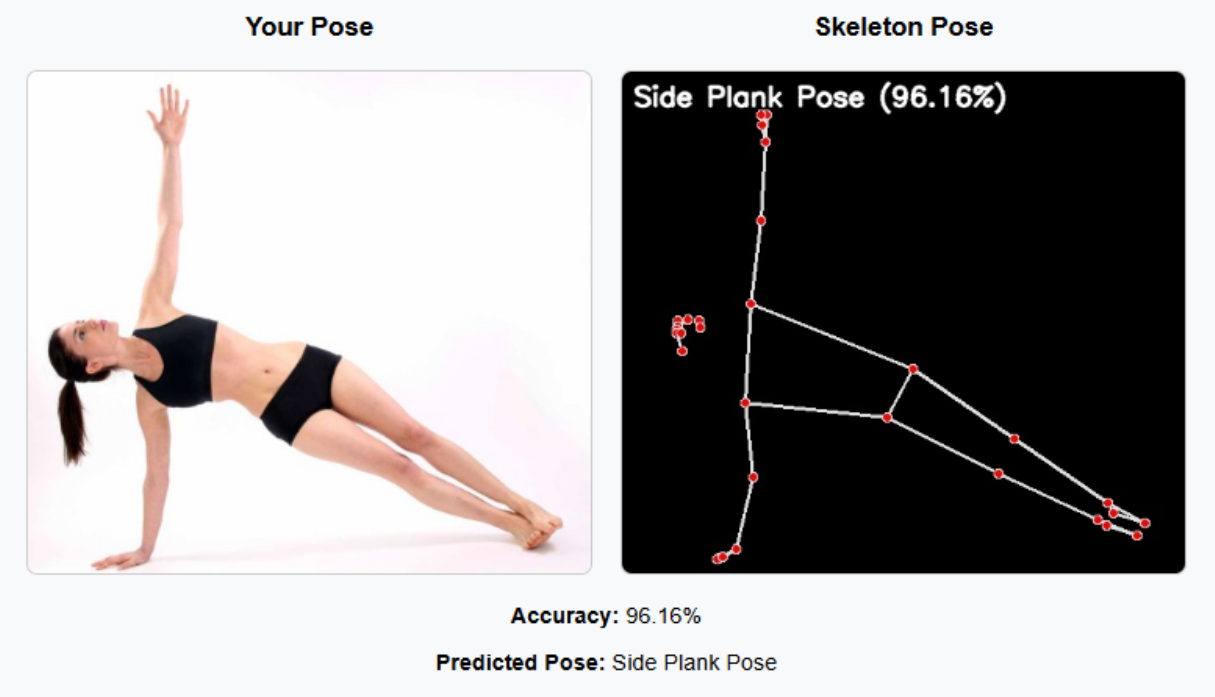}} \\
        \subfloat[\centering]
        {\includegraphics[width=4cm]{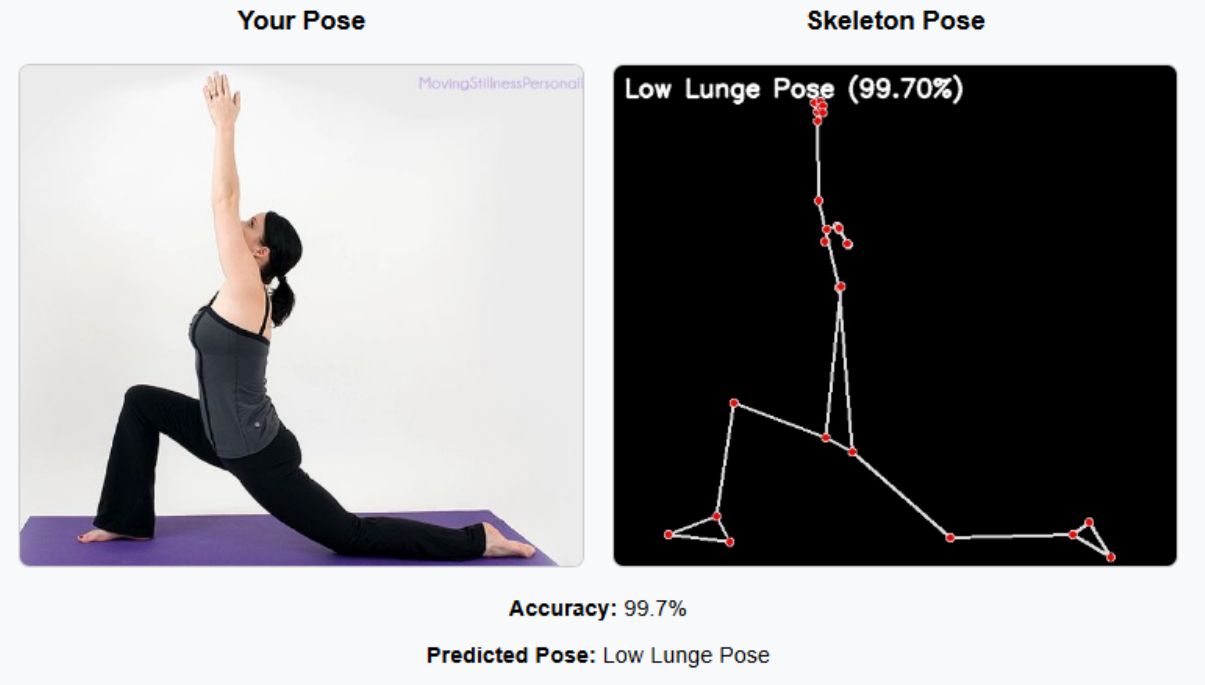}} &

        \subfloat[\centering]
        {\includegraphics[width=4cm]{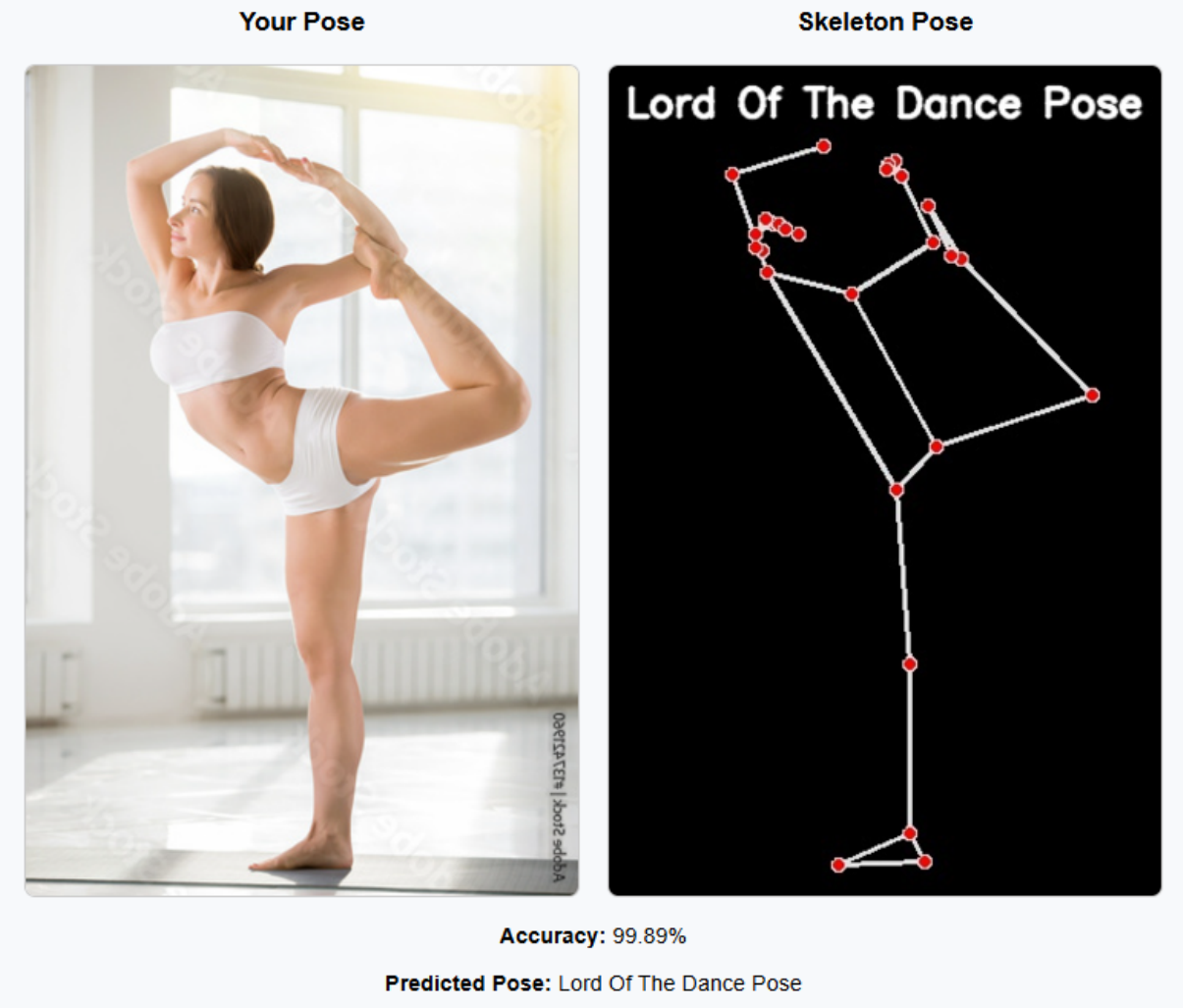}} \\

    \end{tabular}
    \caption{Accurately Predicted Yoga Pose Samples: (a) Goddess Pose (b) Side Plank Pose (c) Low Lunge Pose (d) Lord of the Dance Pose}
    \label{fig:acc_pred}
\end{figure}

\subsection{Critical Analysis and Feature Visualization of the Best Performing Model}
Critical insights into the feature extraction process of the VGG16 model when classifying poses using Mediapipe skeleton input are brought forth by Grad-CAM visualization in Fig. \ref{fig:gradcam_vgg16}. Grad-CAM, which stands for Gradient-weighted Class Activation Mapping, is a technique for visualizing what parts of an input image are most responsible for a model’s classification decisions. In this case, each subfigure represents a distinct yoga pose class, with the highlighted regions (in warmer colors) representing the key areas that the model looks at when making classification decisions.

 {MediaPipe-based skeletons likely outperform YOLOv8-Pose because BlazePose detects 33 keypoints with higher stability and consistency across joint locations, providing more precise skeletal representations that improve VGG16’s ability to discriminate subtle pose variations. Beyond accuracy, MediaPipe offers several advantages: it is lightweight and optimized for real-time inference, making it highly suitable for mobile and embedded devices; it provides anatomically consistent landmark detection across a wide range of body orientations and occlusions; and it includes built-in smoothing and temporal filtering mechanisms, which reduce jitter and noise in skeleton keypoints during dynamic poses. These strengths not only enhance pose classification but also increase robustness in real-world applications such as digital fitness or rehabilitation systems.}
\begin{figure}[htbp]
    \centering
    \begin{tabular}{cc}
        \subfloat[\centering]
        {\includegraphics[width=4cm]{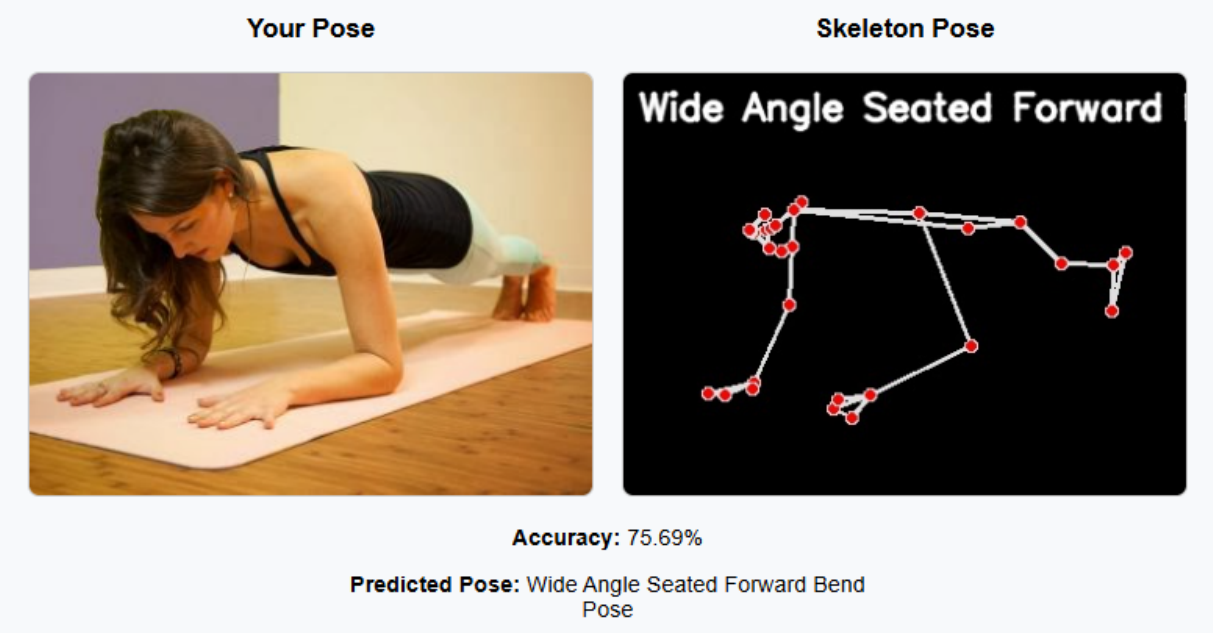}} &

        \subfloat[\centering]
        {\includegraphics[width=4cm]{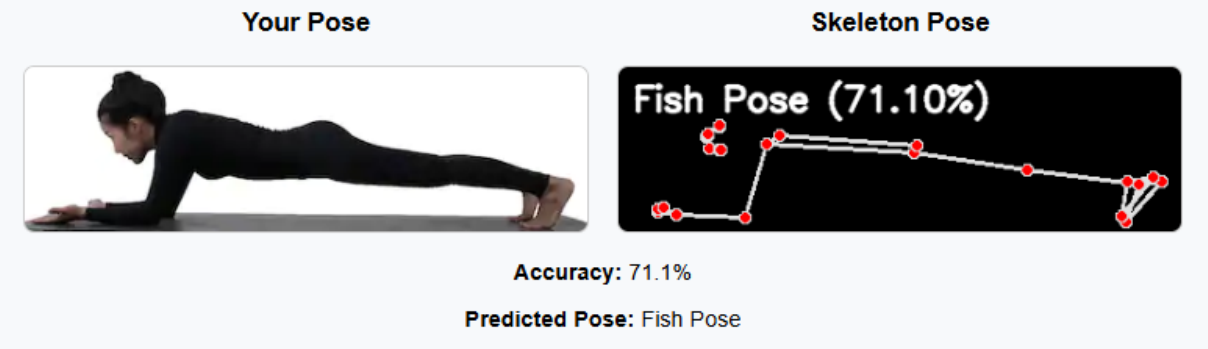}} \\

    \end{tabular}
    \caption{Misclassified Yoga Pose Samples: (a) Actual: Dolphin Plank Pose, Prediction: Wide Angle Seated Forward Bend Pose (b) Actual: Dolphin Plank Pose, Prediction: Fish Pose}
    \label{fig:miss_pred}
\end{figure}

For very different poses like Downward Facing Dog Pose and Warrior II Pose, the model puts strong emphasis on alignments of limbs and angles of joints. The highlighted regions are the locations of important skeletal features, for example, the torso and extended arms that are critical to distinguish between these poses. In the Goddess Pose, the model has realized the symmetrical position of the arms and legs, which is unique to this pose, so it gives itself a confident classification. The visualizations show that the model consistently attends to the most relevant skeletal features for each pose. For example, poses involving intricate configurations of the legs or arms, like the Tree Pose and Lord of the Dance Pose, show focused attention around the corresponding joints and limb configurations. The accuracy of the spotlighted regions suggests that the Mediapipe skeleton input does indeed eliminate noise to a large extent, allowing the VGG16 model to focus on the discriminative features.

The Grad-CAM maps illustrate misclassification prevention, showing the robustness of the model in handling difficult poses like the Side Plank Pose and the Wide-Angle Seated Forward Bend Pose, where overlapping limbs would be problematic for a less skilled model. Through focusing on unique skeletal patterns, the VGG16 model maintains high levels of precision and recall while drastically lowering errors.

The correctly classified poses shown in Fig. \ref{fig:acc_pred} highlight the model’s ability in accurately identifying unique skeletal features associated with poses such as the Goddess Pose and the Lord of the Dance Pose. These types of poses are characterized by marked skeletal structures that are very unique, hence allowing the VGG16 model to distinguish one from another based on these clear contrasts. Furthermore, the incorporation of Mediapipe’s skeletal input presumably facilitated feature extraction, thereby augmenting the model’s resilience in these instances.

On the other side, the misclassifications portrayed in Fig. \ref{fig:miss_pred} are owed to the plain similarities of skeletal structures happening between actual and predicted poses. For example, the misclassification of the Dolphin Plank Pose as the Wide-Angle Seated Forward Bend Pose may indicate the model was misled by either overlapping skeletal keypoints or ambiguous joint angles in its input representation. It can also be that there is not enough discrimination in training data between the two poses—Dolphin Plank and Fish—or such discrimination is not learned, hence the model is ambiguous. These errors underline the difficulty of discrimination of poses with subtle or overlapping skeletal features, which suggests that further refinement of the dataset or model architecture is necessary to improve the sensitivity specific to each pose.

\section{Discussion}\label{sec5}

The Yoga-16 dataset mitigates common limitations, such as class imbalance and low-resolution data, by curating and balancing 16 widely used yoga poses. Combining and augmenting images from public datasets, Yoga-16 provides a robust foundation for evaluating deep learning models. The balanced dataset ensures consistent learning across pose classes, enabling better generalization and reducing biases often affecting classification accuracy.

This study evaluated three input types—direct images, MediaPipe skeletons, and YOLOv8-Pose skeletons—and observed that skeleton-based inputs significantly outperformed direct image inputs. Skeleton representations isolate pose features and remove background noise, enhancing model focus on discriminative features. For instance, MediaPipe skeletons achieved an impressive accuracy of 96.09\% with VGG16, compared to 86.33\% for direct image inputs. This emphasizes the potential of skeleton-based representations in improving classification performance.

However, challenges in skeleton-based representations, such as inconsistencies in landmark extraction between MediaPipe and YOLOv8-Pose, were observed. These inconsistencies led to subtle misclassifications in poses with overlapping or ambiguous joint structures. MediaPipe consistently outperformed YOLOv8-Pose because it provides more stable and anatomically consistent landmarks, particularly for symmetrical poses and fine-grained limb orientations. In contrast, YOLOv8-Pose often struggled with precise keypoint localization in occluded or overlapping regions, resulting in noisier skeleton inputs. Preprocessing techniques and model optimizations were essential to manage these challenges, demonstrating the need for careful refinement in data preparation.

Among the three deep learning architectures evaluated (VGG16, ResNet50, and Xception), VGG16 consistently emerged as the most reliable model across all input types. Its superior performance can be attributed to its effective feature extraction capabilities, particularly when processing skeleton-based inputs. Grad-CAM visualizations revealed that VGG16 successfully focused on critical skeletal features, such as limb alignments and joint angles, to differentiate between poses.

For example, in poses like the Goddess Pose and Warrior II Pose, the model emphasized symmetrical arm and leg alignments and distinctive joint angles, enabling accurate classification. Grad-CAM maps also illustrated the model's robustness in challenging scenarios, such as distinguishing between the Side Plank Pose and Wide-Angle Seated Forward Bend Pose, where overlapping limbs might confuse less advanced models.

Despite its high accuracy, the model faced difficulty classifying poses with subtle or overlapping skeletal features, such as Dolphin Plank Pose and Wide-Angle Seated Forward Bend Pose. These misclassifications stemmed from overlapping skeletal key points or insufficient discrimination in the training data. This highlights the need to refine the dataset and model architecture further to enhance sensitivity to subtle pose differences.

Additionally, the Grad-CAM analysis showed that the model occasionally failed to focus on uniquely defining skeletal patterns in similar poses, underlining the importance of feature diversity and pose-specific data augmentation during training.

The results highlight the utility of skeleton-based representations in reducing noise, isolating pose-specific features, and achieving high accuracy in yoga pose classification. With balanced datasets and effective architectures like VGG16, future systems can offer reliable real-time feedback, advancing digital fitness technologies. Furthermore, the strong generalization capability demonstrated on diverse real-world data underscores their practical applicability in real-world scenarios.

\section{Conclusion and Future Work}\label{sec6}

This study establishes a solid foundation for automated yoga pose classification, emphasizing the significance of skeleton-based representations and the integration of robust neural network architectures. By introducing the Yoga-16 dataset and achieving remarkable classification accuracy, particularly with the VGG16 model and MediaPipe skeleton inputs, it demonstrates the potential of leveraging structured pose data to enhance precision and efficiency in digital fitness applications. Additionally, we evaluated the model's generalization capability using a custom dataset derived from diverse real-world scenarios.

 {However, the relatively limited size of the Yoga-16 dataset raises concerns regarding generalization. We have implemented cross validation to mitigate the concern but increasing the size of the dataset will make the system more reliable without any concerns. Another limitation arises from the difficulty of distinguishing poses with subtle or overlapping skeletal features, such as Dolphin Plank and Wide-Angle Seated Forward Bend, where ambiguities in joint positions can mislead the model. Future work should focus on larger and more diverse datasets, real-time pose recognition, multimodal input representations, and improved model interpretability to address these challenges.} Integrating AR/VR technologies could provide immersive, interactive feedback, boosting user engagement and accessibility. Expanding the dataset and refining model sensitivity to ambiguous skeletal patterns would further improve robustness and generalizability, enabling automated yoga pose classification systems to evolve into comprehensive tools for personalized instruction, promoting wellness and accessibility across diverse populations.

\section*{Funding Declaration}\label{sec7}
Not applicable.

\section*{Declaration of Competing Interest}\label{sec8}
The authors declare no conflicts of interest related to this study.

\section*{Availability of Data and Materials}\label{sec9}
All the data and materials have been included in the submission. The data are available at  \url{https://www.kaggle.com/datasets/mohiuddin2531/yoga-16/data} by \cite{b34} and the codes are available at \url{https://github.com/mohiuddin2531/yoga-16} by \cite{b37}.
\section*{Ethical Consideration}\label{sec12}

The Yoga-16 dataset was created by choosing and merging images from the Yoga-82 dataset \cite{b19}, and the Yoga Poses Dataset \cite{b33}. To achieve ethical compliance, the following measures were implemented:
\subsection*{Data Source, Privacy and Authorization}
Yoga-16 was created using the Yoga-82 and Yoga Poses Dataset, all freely available. Their inclusion followed the various licensing terms and usage conditions, assuring conformity with their sources' academic and non-commercial aims. The dataset excludes identifiable personal information, with faces masked or not the primary emphasis, reducing privacy hazards. The generalization dataset \cite{b36} assures that the videos are publicly accessible, not restricted by privacy settings, free of copyright claims, and utilized for research purposes following fair use guidelines. It also ensures correct attribution and records the selection process for transparency.

\subsection*{Ethical Approval}

The Ethics Committee at Premier University recognized that this work did not require ethical assessment because it only used publicly available, non-identifiable datasets.




\newpage

\end{document}